\documentclass{article} 
\PassOptionsToPackage{table}{xcolor} 
\usepackage[final]{colm2026_conference}

\usepackage{microtype}
\usepackage{hyperref}
\usepackage{url}
\usepackage{booktabs}
\usepackage[T1]{fontenc}

\usepackage{lineno}

\usepackage{hyperref}
\usepackage{url}
\usepackage{booktabs}
\usepackage{amsmath}
\usepackage{graphicx}
\usepackage{lineno}
\usepackage[table]{xcolor}
\usepackage{mathtools}
\usepackage{makecell}
\usepackage{algorithm}
\usepackage{algorithmicx}
\usepackage[noend]{algpseudocode}
\usepackage{wrapfig}
\usepackage{float}
\usepackage{comment}
\usepackage{multirow}
\usepackage{amssymb}
\usepackage{arydshln}
\usepackage{subcaption}
\usepackage{enumitem}
\usepackage{arydshln}
\usepackage{tabularx}
\usepackage{wasysym}
\usepackage{colortbl}
\usepackage{pifont}
\definecolor{lightyellow}{RGB}{209, 209, 209} 

\newcommand{\graymidrule}{\arrayrulecolor{tablelightgray}\midrule\arrayrulecolor{black}}

\DeclarePairedDelimiter\floor{\lfloor}{\rfloor}

\definecolor{embeddingcolor}{RGB}{25, 2, 173} 
\definecolor{summarizercolor}{RGB}{138, 3, 3} 
\definecolor{algcolor}{RGB}{1, 92, 7} 
\definecolor{embeddingcolorlight}{RGB}{232, 237, 255} 
\definecolor{summarizercolorlight}{RGB}{255, 214, 214} 

\definecolor{darkblue}{rgb}{0, 0, 0.5}
\hypersetup{colorlinks=true, citecolor=darkblue, linkcolor=darkblue, urlcolor=darkblue}

\def\mystrut(#1,#2){\vrule height #1pt depth #2pt width 0pt}   

\definecolor{purple}{rgb}{0.5,0,1}
\definecolor{dcyan}{rgb}{0.2,0.6,0.5}
\definecolor{light-gray}{gray}{0.95} 

\definecolor{darkgreen}{RGB}{0,140,0}
\definecolor{darkred}{RGB}{200,0,0}
\definecolor{lightgreen}{RGB}{189,252,192}
\definecolor{lightred}{RGB}{255,205,212}
\definecolor{lightyellow}{RGB}{255,240,160}
\definecolor{lightblue}{RGB}{195,221,255}
\definecolor{lightpurple}{RGB}{232,209,255}

\usepackage{colortbl}
\usepackage{listings}
\usepackage{array}
\newcommand{\PreserveBackslash}[1]{\let\temp=\\#1\let\\=\temp}
\newcolumntype{C}[1]{>{\PreserveBackslash\centering}p{#1}}
\newcolumntype{R}[1]{>{\PreserveBackslash\raggedleft}p{#1}}
\newcolumntype{L}[1]{>{\PreserveBackslash\raggedright}p{#1}}

\newcommand{\nameAlg}[0]{\textsc{Scychic}}

\definecolor{kh}{HTML}{168aff}

\usepackage{mathtools}

\definecolor{lightblue}{RGB}{235,244,250}
\definecolor{lightgreen}{RGB}{237,246,240}
\definecolor{lightpurple}{RGB}{242,240,247}

\definecolor{codegreen}{rgb}{0,0.6,0}
\definecolor{codegray}{rgb}{0.5,0.5,0.5}
\definecolor{codepurple}{rgb}{0.58,0,0.82}
\definecolor{backcolour}{rgb}{0.95,0.95,0.92}

\definecolor{tablelightgray}{RGB}{255, 245, 217}
\title{From Papers to Panoramas: \\ Building Hierarchies of Scientific Literature at Scale}

\author{
Muhan Gao$^{1*}$, \ Jash Shah$^{2}$, \ Weiqi Wang$^{2}$, \ Kuan-Hao Huang$^{1}$, \ Daniel Khashabi$^{2}$ \\[4pt]
$^{1}$Texas A\&M University, \ $^{2}$Johns Hopkins University
}

\begin{document}
\ifcolmsubmission
\linenumbers
\fi

\maketitle

\begin{abstract}
Scientific knowledge is growing rapidly, making it difficult to track progress 
and high-level conceptual links across broad disciplines. 
 While tools like citation networks and search engines help retrieve related papers, they lack the abstraction needed to capture the \textit{density} and structure of activity across subfields.
We motivate the goal of 
 organizing broad swaths of scientific literature into a high-quality hierarchical structure that spans multiple levels of abstraction---from broad domains to specific studies. 
Such a representation can provide insights into which fields are well-explored and which are  under-explored. 
To achieve this goal, we develop a hybrid approach that combines efficient embedding-based clustering with LLM-based prompting, striking a balance between \textit{scalability} and \textit{semantic precision}. Compared to LLM-heavy methods like iterative tree construction, our approach achieves superior quality-speed trade-offs.
Our hierarchies capture different dimensions of research contributions, reflecting the interdisciplinary and multifaceted nature of modern science. We evaluate its utility by measuring how effectively an LLM-based agent can navigate the hierarchy to locate target papers. Results show that our method improves interpretability  and offers an alternative pathway for exploring scientific literature beyond traditional search methods.\footnote{
\href{https://github.com/JHU-CLSP/science-hierarchography}{Code and demo are available online}. * Work done at JHU.
}

\end{abstract}

\section{Introduction}
The pace of scientific publishing is accelerating, but this growth is uneven across fields~\citep{hope2023computational}. Some areas attract dense research activity, while others remain underexplored. This raises a natural question:
\begin{center}
\textit{How do we understand the \underline{distribution} of scientific efforts across different sub-areas?}
\end{center}
Answering this question is essential \textit{for both academic and policy stakeholders}. A clearer view of how research efforts are distributed enables institutions to spot emerging or neglected areas, prioritize strategic hiring and future  agendas. For policymakers, it supports more informed funding decisions, ensuring that critical but underexplored domains receive the attention and resources they deserve.



Conventional tools like Google Scholar are designed as retrieval engines, optimized to return a handful of papers that match a specific query. They offer little in the way of a comprehensive or structured view of the broader scientific landscape.  
Similarly, while modern LLM-based assistants can surface related works (seen during pretraining or via their retrieval tools), they fall short in offering a broad, bird’s-eye perspective on scientific progress.


Addressing this challenge requires abstraction: a way to generalize over research problems and techniques and to connect broad scientific areas to specific papers via intermediate categories. At one end, we have high-level domains (e.g., physics, AI); at the other, individual papers. Between them lies a latent spectrum of subfields and methodological clusters. What’s missing is a data structure that captures all these abstraction levels.

We propose building large-scale hierarchical representations of scientific literature. A well-designed hierarchy provides a macro-level view of scientific progress, revealing how research is distributed across methods and application areas. This helps researchers spot emerging trends and gaps, and supports policymakers and institutions in making more strategic resource decisions. It also offers a new way to explore the literature---complementing traditional search by allowing users to navigate science through conceptual hierarchies. 



\begin{wrapfigure}[21]{r}{0.6\textwidth}
    \centering
    \vspace{-0.6cm}
    \includegraphics[width=0.95\linewidth, trim=6.92cm 0.500cm 1.6cm 1.2cm, clip=true]{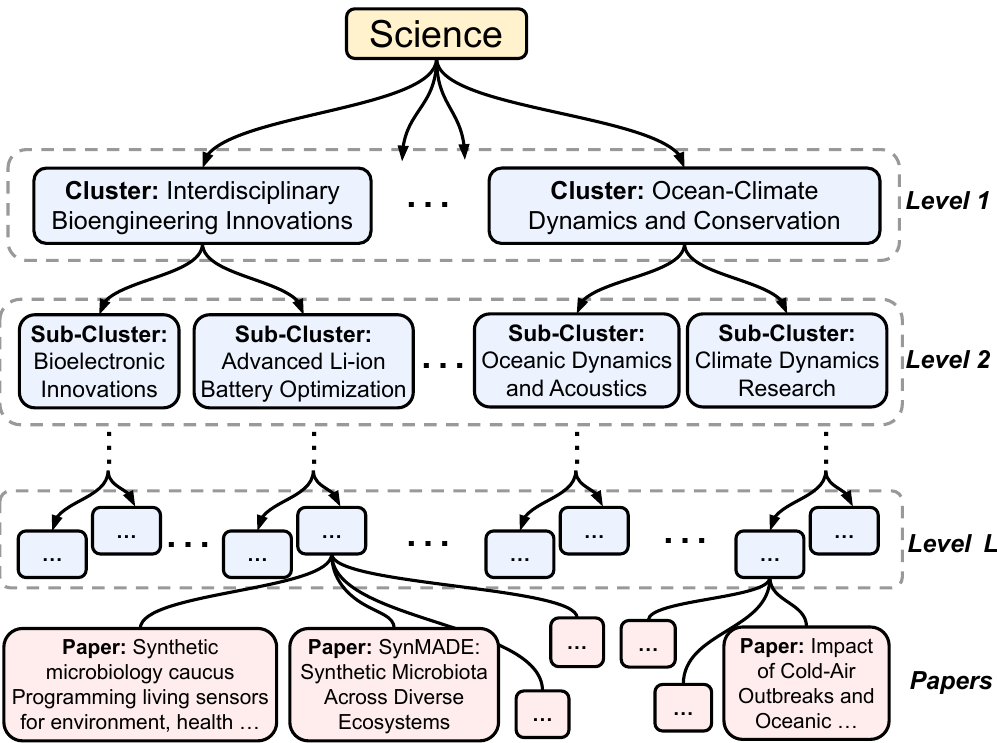}
    \caption{
    An illustration of how scholarly work can be organized hierarchically, from broad research domains at the top, through increasingly specific sub-clusters, down to papers at the lowest level. Critically, this structure must be inferred automatically and at scale.
    }
    \vspace{-0.3cm}
    \label{fig:teaser}
\end{wrapfigure}

\textbf{How should scholarly work be represented?}
A central challenge in building a scientific hierarchy is defining what each node represents. Research papers often span multiple topics (e.g., \textit{reinforcement learning for medical imaging} or \textit{deep learning for oceanography}). To capture this complexity, we develop a prompting strategy that decomposes papers into key \textit{contribution} types---such as the \textit{problems addressed} and \textit{techniques used} (\S\ref{subsec:representing:papers}). For each fixed contribution type, we construct a corresponding hierarchical structure, ensuring that papers are organized into meaningful, coherent categories.


\textbf{What construction strategies balance scalability and quality?}
To address this, we introduce \nameAlg{} (pronounced ``psychic''), a new method for building high-quality hierarchical structures of scientific literature. \nameAlg{} integrates fast embedding-based clustering with LLM prompting, combining the efficiency of embeddings with the semantic precision of LLMs (\S\ref{sec:muhan:approach}).


\textbf{How can we evaluate the quality of a scientific hierarchy?}
Scientific hierarchies lack a fixed ground truth---they evolve over time as research landscapes shift. We therefore adopt an \textit{evaluation-through-utilization} approach, measuring \textit{whether an information seeker (human or AI) can efficiently locate specific content} (e.g., child nodes) by navigating the hierarchy from the root. This evaluation hinges on the idea that a good hierarchy enables rapid information discovery, even though its utility extends well beyond search alone (\S\ref{subsec:evaluation:utiliization}).


\textbf{What did our empirical results show?}
Our approach achieves the best trade-off between quality and speed when compared to LLM-heavy methods like iterative tree construction. Extensive experiments show that \nameAlg{} consistently produces higher-quality hierarchies than a broad set of baselines (\S\ref{sec:findings}).  Validation on a 10K-paper dataset further confirms its strong accuracy and scalability for large-scale use.


\textbf{Contributions:}
(1) We introduce the goal of constructing large-scale, abstract hierarchies of scientific literature to reveal how scholarly efforts are distributed across research areas.
(2) We propose a utilization-based evaluation framework that measures how effectively users can discover information by traversing the hierarchy.
(3) We present \nameAlg{}, a new method that combines fast embedding-based clustering with LLM prompting to build high-quality, multidimensional hierarchies. Extensive experiments show that \nameAlg{} outperforms baseline approaches, offering a more structured and bird's-eye view of scientific progress.

\definecolor{highlight}{RGB}{192,0,0} 
\newcommand{\cmark}{\ding{51}}
\newcommand{\xmark}{\ding{55}}
\newcommand{\richbar}[2]{\textcolor{#1}{\rule{#2}{5pt}}}
\definecolor{highlight_green}{RGB}{0,0,0}

\begin{table}[ht]
    \fontsize{8pt}{9pt}\selectfont 
    \setlength{\tabcolsep}{8pt}
    \renewcommand{\arraystretch}{1.1}
    \centering
    \resizebox{\linewidth}{!}{%
        \begin{tabular}{L{3.5cm}L{1.8cm}L{5cm}cc}
            \toprule
            System & Depth & Node description & Auto & Public \\
            \midrule
            \href{https://arxiv.org/category_taxonomy}{arXiv Taxonomy}
              & Two & \richbar{gray}{8pt}\; keyword & \xmark & \cmark \\
            \href{https://www.scopus.com/home.uri}{Scopus} 
              & Three & \richbar{gray}{8pt}\; keyword & \xmark & \cmark \\
            \href{https://openalex.org/}{OpenAlex} 
              & Four & \richbar{gray}{16pt}\; keywords & \cmark & \cmark \\
            \href{https://pubmed.ncbi.nlm.nih.gov/}{PubMed MeSH} 
              & Multiple & \richbar{gray}{8pt}\; medical term & \xmark & \cmark \\
            \href{https://www.microsoft.com/en-us/research/project/microsoft-academic-graph/}{Microsoft Academic Graph} 
              & Multiple & \richbar{gray}{16pt}\; keywords & \cmark & \dag \\
            \midrule
            \rowcolor{teal!6}
            \textbf{Ours}
              & Multiple & \richbar{teal}{50pt}\; detailed summary & \cmark & \cmark \\
            \bottomrule
        \end{tabular}
    }%
    \caption{
        Comparison of hierarchical resources for organizing scientific literature, ordered by hierarchy depth.
        Conventional systems are built for indexing, relying on fixed, shallow taxonomies with keyword-based nodes and human-assigned labels. In contrast, our method supports deeper, designer-controlled hierarchies with rich natural-language summaries, enabling more flexible and exploratory analysis of scientific work.
        \cmark\ = yes/automated, \xmark\ = no/manual.
        \dag\ Discontinued in 2021.
    }
    \label{tab:system_comparison}
    \vspace{-2.5em}
\end{table}

\section{Related Work}
\label{sec:related:work}

\textbf{Taxonomy induction:} The field of taxonomy induction has progressed from early pattern-based techniques~\citep{hearst-1992-automatic,pantel2006espresso,yang-callan-2009-metric,girju2006automatic} to modern LLM-augmented methods. \citet{wan2024tnt,zeng2024chain,chen2023prompting,DBLP:journals/corr/abs-2408-09070} apply zero-/few-shot reasoning and ensemble ranking, while others explore open-ended, vocabulary-free taxonomy creation~\citep{gunn2024creating}, self-supervised expansion in low-resource domains~\citep{mishra2024flame}, and graph-based methods leveraging metadata and citations~\citep{cong2024openforgeprobabilisticmetadataintegration,sas2024automaticbottomuptaxonomyconstruction,shen2024unifiedtaxonomyguidedinstructiontuning}. Optimization and in-context learning  have also shown promise~\citep{DBLP:journals/corr/abs-2410-03761,DBLP:conf/www/Shi0CW024,xu2025compress,DBLP:conf/starsem/JainA22,DBLP:conf/naacl/ChenLK21}.

A common characteristic of the methods above is that they operate at the entity or term level: each node in the taxonomy is a keyword or short phrase (e.g., "reinforcement learning," "BERT"). TaxoGen~\citep{zhang2018taxogen} extracts term-level hierarchies by clustering frequent noun phrases from text corpora; however, when applied to academic abstracts, it tends to surface generic terms (e.g., ``new method,'' ``novel framework'') rather than meaningful concepts suitable for organizing scholarly work. Chain-of-Layer~\citep{zeng2024chain} demonstrates strong performance on entity taxonomies but exhibits degradation beyond 80 entities, highlighting the scalability challenges of prompt-heavy approaches.

In contrast, our goal is to organize broad swaths of scientific literature into navigable hierarchies that reveal how research efforts are distributed across subfields, requiring a scale of thousands of papers with rich multi-sentence node descriptions across diverse domains.

Beyond term-level taxonomies, several prior works address the organization of scientific knowledge at the topic or paper level. HiGTL~\citep{hu2025taxonomytree} constructs topic-level concept hierarchies from citation graphs with supervised joint optimization, and Sci-OG~\citep{pisu2025hybrid} classifies pairwise relationships between research topics using labeled triples and corpus statistics. Both produce topic-level rather than paper-level structures, and rely on external supervision not available for arbitrary paper collections. \citet{zhu2025context} proposes LLM-guided multi-aspect clustering for survey-based collections of fewer than 100 papers. \citet{oarga2024scientific} builds domain-specific hierarchies (e.g., Chemical) using iterative LLM refinement, operating at a small scale within a single domain. TaxoAdapt~\citep{kargupta-etal-2025-taxoadapt} iteratively expands LLM-generated taxonomies into DAG-structured hierarchies spanning multiple dimensions, evaluated on collections of up to a few thousand papers; we compare with it empirically in \S\ref{sec:compare-alg}. Our work is, to our knowledge, the first to construct multi-level hierarchies over thousands of papers across diverse scientific domains, without requiring citation graphs, ground-truth taxonomies, or labeled relationship data.

\noindent
\textbf{Structured representation of science:}
As science grows at an unprecedented rate~\citep{teufel1999annotation, pertsas2017scholarly, constantin2016document, fisas2016multi, liakata2010corpora}, numerous frameworks have emerged to structure this information through knowledge graphs and taxonomies~\citep{fathalla2017towards, jaradeh2019open, oelen2020generate, vogt2020toward, soldatova2006ontology}. Recent work includes prompt-based topic modeling~\citep{pham2023topicgpt}, iterative taxonomy construction that incorporates object properties and graph mining~\citep{cui2024automated, marchenko2024taxorankconstruct}, and hybrid approaches that combine curated ontologies with data-driven maps~\citep{zimmermann2024ontoverse}.
A complementary line of work quantifies research influence, e.g., the SciImpact benchmark~\citep{zhu-etal-2026-sciimpact}, whereas our hierarchies surface how research \textit{effort} is distributed across subfields.
Our work builds on these efforts by constructing a high-quality hierarchical structure tailored to scientific literature, in three key ways. The prior work: 
(1) Produces shallow hierarchies, typically only one or two levels deep;
(2) Uses cluster labels based on keywords, whereas ours are derived from natural language summaries of papers;
(3) Depends heavily on manual effort, while our pipeline is fully automated. 

\paragraph{Structured knowledge in LLMs:}
Prior work has explored how LLMs internalize hierarchical knowledge.
For example, \cite{he2024language,melgarejo2023probing,park2025the} extend the linear representation hypothesis to reveal that LLMs encode categorical concepts as polytopes, with hierarchical relationships reflected as orthogonal directions.
Others examine explicit hierarchical structures and prompting for guiding LLMs~\citep{wolfman-etal-2024-hierarchical,budagam2024prompting}, or materialize large-scale structured knowledge bases~\citep{moskvoretskii-etal-2024-large,hu2024gptkbcomprehensivelymaterializingfactual}.
In line with the same aspirations, our work explores the use of hierarchical structures to organize scientific literature.

In Table~\ref{tab:system_comparison} we summarize the  differences with existing hierarchical resources. Most prior systems are limited to fixed depth and rely on manually assigned labels for indexing---a process often prone to bias~\citep{hadfield2020delay}.
A detailed comparison with existing systems is in \S\ref{appendix:relate_work}.

\section{From Papers to Panoramas: Towards Hierarchy of Scholarly Work}






We define the problem (\S\ref{subsec:definition}), scholarly contributions (\S\ref{subsec:representing:papers}), and hierarchy depth (\S\ref{subsec:depth}).


\subsection{Formal Problem Statement}
\label{subsec:definition}

We define the task as an inference problem where the \textbf{input} is a large set of scientific papers: \(P = \{p_1, p_2, \ldots, p_n\}\). 
The goal is to infer a \textit{hierarchical structure} (i.e., a tree) 
for a specific contribution type (e.g., problem statement) of a collection of papers. 
The nodes of this tree are the atomic concepts representing scholarly ideas or goals.
The edges  (relations connecting two nodes) 
encode whether one node is a specific version of another node (i.e., ``isA'' relationships) which  defines a hierarchical link between node pairs, indicating a child node is a subclass of its more abstract parent node (e.g., ``RLHF isA RL'' means ``RLHF'' is a type of ``RL''). 
The specific papers $P$ are the nodes of this tree. 
The overall hierarchy represents levels of specificity and abstraction, with nodes closer to the root representing broader topics. Broader topics are at the upper levels, while more specific subtopics and individual papers are at the lower levels.

\noindent
\textbf{Why a tree structure?} 
A tree offers a clear, interpretable way to capture hierarchical relations among scientific ideas, showing how concepts specialize or generalize without cycles or ambiguity. 
Each node inherits meaning from its ancestors, tracing the progression from broad themes to concrete contributions.
We build \textit{one tree per contribution type} (e.g., problem, method; \S\ref{subsec:representing:papers}), forming a \textbf{forest of hierarchies} that reflects the multi-faceted and interdisciplinary nature of scientific research.

\subsection{Decomposing Papers to Contributions}
\label{subsec:representing:papers}


A central challenge is how to represent the content of scholarly work within hierarchy nodes. Scientific papers are idea-dense, often combining broad goals, specific problems, and technical methods. To capture this complexity, we extract structured representations that disentangle these distinct aspects~\citep{d2020nlpcontributions}. This also mitigates the issue of input length: papers typically range from 4 to 10 pages (5K to 10K tokens), making full-document processing across large corpora infeasible and costly for LLMs.

We use an LLM (\texttt{gpt-4o}) to preprocess each paper (title and abstract) and request the LLM to break them down into a \textbf{pre-defined set of contributions}, akin to prior work \citep{hope2017accelerating,chan2018solvent} that mines ``problem schema'' from existing documents. We consider the following contribution types: (1) \textbf{problem statement} (the problem addressed), (2) \textbf{solution} (the technical approach used), (3) \textbf{result} (the key finding), and (4) \textbf{topic} (the overarching themes). 
(See \S\ref{appendix:contributions} for prompts and examples).
We note that each contribution may include additional dimensions (sub-contributions). For instance, a ``result'' encompasses both the ``outcome'' and its ``potential impact.'' In total, this yields $C=11$ sub-contributions per paper. 





\vspace{-0.3cm}
\subsection{Choosing a Hierarchy Depth}
\label{subsec:depth}
\vspace{-0.2cm}
While the ideal number of hierarchy layers is ultimately empirical, we can build useful intuition from the structure of a near-balanced tree. For a tree with branching factor $b$ and depth $L$, the total number of nodes is roughly $O(b^L)$. To organize $C$ contributions, the number of nodes should scale with $C$, implying a depth of $L = O(\log_b C)$. In practice, we use $L = 3$ for a 2K-paper corpus and $L = 4$ for 10K papers, consistent with this logarithmic scaling. Extrapolating further, corpora of $10^7$ papers would likely require depths of $L = 6$ or $7$.
The layer-wise cluster schedules used in our experiments, together with an empirical sensitivity analysis, are reported in \S\ref{subsection:additional_ana}.


\vspace{-1em}
\section{Method for Constructing Scientific Hierarchies}
\label{sec:approaches}

We present algorithms to address our proposed goal. 
We start with our main method, \nameAlg{} (\S\ref{sec:muhan:approach}),  and discuss its special cases (\S\ref{sec:baseline:topdown:bottomup}).

\newcommand{\embedder}{\texttt{\textcolor{embeddingcolor}{embedder}}}
\newcommand{\summarizer}{\texttt{\textcolor{summarizercolor}{summarizer}}}
\newcommand{\alg}{\texttt{\textcolor{algcolor}{clusterer}}}

\definecolor{darkgreen}{RGB}{0,140,0}

\newcommand{\Mycomment}[1]{\Comment{\textcolor{darkgreen}{#1}}}

\begin{wrapfigure}[22]{R}{0.56\linewidth}
      \vspace{-0.8cm}
      \begin{minipage}{\linewidth}
\begin{algorithm}[H]
  \fontsize{8}{10}\selectfont 
  \caption{\nameAlg{} algorithm}
  \label{alg:namealg}
  \begin{algorithmic}[1]
    \Require{Set of papers $P = \{p_1, p_2, \ldots, p_n\}$, \embedder, \alg, \summarizer, num of layers $L$, target cluster sizes $(k_1, k_2, \ldots, k_L)$}
    \State \textbf{Initialization:} For each paper $p_i \in P$, use \embedder{} to embed its selected components and concatenate them into a single embedding in $\mathbb{R}^{d \cdot |C'|}$.

    \For{layer $l = 1$ to $\lfloor L/2 \rfloor$} \Mycomment{Top-down phase}
        \If{$l = 1$}
            \State Apply \alg{} to divide papers into $k_1$ clusters
        \Else
            \For{each cluster from layer $l-1$}
                \State Apply \alg{} to divide into subclusters
            \EndFor
        \EndIf
        \State Use \summarizer{} to generate summaries for clusters
    \EndFor
    \For{each cluster $\tau$ at level $\lfloor L/2 \rfloor$} \Mycomment{Bottom-up phase} 
        \For{layer $l = L$ to $\lfloor L/2 \rfloor + 1$}
            \If{$l = L$}
                \State Collect the embeddings of papers within $\tau$. 
            \Else
                \State Apply \embedder{} on  summaries of cluster $l+1$  
            \EndIf
            \State Apply \alg{} to form higher-level clusters
            \State Use \summarizer{} to generate summaries
        \EndFor
    \EndFor
    \State \textbf{return} Hierarchical structure
  \end{algorithmic}
\end{algorithm}
\vspace{-1em}
\end{minipage}
\end{wrapfigure}

While all approaches leverage LLMs to some extent, they differ significantly in their reliance on them: some require many calls (linear or quadratic in the number of papers), while others are more efficient (e.g., logarithmic). Since our goal is to scale to a large number of papers, minimizing LLM usage is critical. Our objective is to identify a method that yields the highest-quality hierarchy with the lowest LLM overhead, balancing quality, latency, and cost.


\subsection{\nameAlg: Alternating Between Clustering and Summarization}
\label{sec:muhan:approach}

\noindent
\textbf{Overview:} Our method builds each contribution-type hierarchy through two complementary stages: a \textit{top-down} phase that clusters paper embeddings into progressively finer subgroups and summarizes each cluster, followed by a \textit{bottom-up} phase that embeds and reclusters the generated summaries to form higher-level abstractions. The combined process yields coherent, interpretable hierarchies that capture both fine-grained  and global structure.

\noindent
\textbf{Ingredients:}
This approach is based on the following design choices:
(1) access to \embedder, a neural model that converts a description into a $d$-dimensional vector, (ideally) capturing its semantic meaning;
(2) a clustering algorithm \alg{} that, given the hyperparameter $k$, generates $k$ clusters;
(3) a contribution type (e.g., problem definition) and its dimensions $C'$ 
extracted per paper as detailed in \S\ref{subsec:representing:papers}  which determines the focus of the node descriptions;
(4) \summarizer, an LLM that generates a summary description which (ideally) provides a more abstract description of a collection of node descriptions; and
(5) the total number of hierarchy layers $L$ and target number of clusters in each layer $(k_1, k_2, \hdots, k_L)$.

Specifically, for  \embedder{} we use \texttt{gte-Qwen2-7B-instruct}, 
for our \summarizer{} we use Llama-3.3-70B-Instruct~\citep{grattafiori2024llama3herdmodels}, and for \alg{}, we apply k-means clustering. (further details in \S\ref{appendix:hyperparams}.)

\noindent
\textbf{Initialization:} The approach begins by embedding each paper.  For each paper $p_i$,  we embed each component in $C'$: $\embedder(c_j^i) \in \mathbb{R}^d$, where $j \in C'$. 
This process results in $|C'|$ component embeddings per paper, which we concatenate into a single embedding in $\mathbb{R}^{d \cdot |C'|}$.
We now present the main algorithm consisting of two phases:

\noindent
\textbf{Phase 1: Top-down:} 
We begin with a \textbf{top-down} strategy that recursively partitions the paper set through the upper half of the hierarchy ($l \in [1, \floor*{L/2}]$). At the first level, all papers are clustered into $k_1$ groups using their embeddings. Each cluster is then processed independently—papers within a cluster are reclustered using \alg{} to form finer subgroups. The number of subclusters assigned to each parent cluster scales linearly with its paper count, ensuring denser regions of the corpus receive finer resolution. This recursive subdivision continues until level $\floor*{L/2}$, producing a coarse-to-fine hierarchy. 
At each level, \summarizer{} generates abstracted summaries for each of the clusters based on the clustered papers' titles and abstracts. The generated cluster description follows the same structure or style as the input descriptions. For example, if the inputs are statements about problem categories, the summaries are also in the same style, but more abstract.

\noindent
\textbf{Phase 2: Bottom-up:} 
We switch to a \textbf{bottom-up} strategy to construct the remaining levels ($\floor*{L/2}+1$ through $L$). 
To form clusters for bottom-level (layer $L$), 
we apply \alg{} to the paper embeddings within each sub-cluster within  level-$\floor*{L/2}$ (the lowest level clustering obtained from the top-down approach). 
We then use the \summarizer{} to create an abstracted description for each cluster.
We repeat this process for all layers from $L$ to  $\floor*{L/2}+1$. 
To build layer $l$, we start by embedding the generated cluster summaries from the level below $l-1$ using \embedder{}, similar to how we embedded the papers. We then run \alg{} on these new embeddings and generate abstracted summaries for the clusters to group these summaries into higher-level clusters. This bottom-up aggregation continues until we connect with the previously constructed level $\floor*{L/2}$ clusters.

\noindent
\textbf{Rationale behind the hybrid design:}
The hybrid approach merges the strengths of top-down and bottom-up strategies. A bottom-up method may create less coherent top-level clusters. 
The top-down approach ensures high-quality top-level clusters but doesn't utilize the abstracted summaries from \summarizer{} used by bottom-up clustering. By combining both methods, the hybrid design achieves robust and effective clustering. Our empirical results in \S\ref{sec:findings} demonstrate this approach's strength by balancing quality and scalability.




\newcommand{\baseline}{\textsc{fLMSci}}
\newcommand{\baselineOne}{\textsc{fLMSci} (parallel)}
\newcommand{\baselineTwo}{\textsc{fLMSci} (incremental)}

\subsection{Computational Complexity}  

\begin{wraptable}[6]{r}{5.1cm}
\setlength{\tabcolsep}{0.5pt}
\small
\centering
\vspace{-0.44cm}
\begin{tabular}{cc}
\toprule  
Approach & \# of LLM calls \\
\midrule
\nameAlg{} & $O\left(C/b\right)$ \\  
\baselineOne{} & $O(C / l)$\\  
\baselineTwo{} & $O(C \log_b C)$\\  
\bottomrule
\end{tabular}
\vspace{-0.1cm}
\vspace{-1em}
\label{tab:complexity}
\end{wraptable} 
A major scalability bottleneck in hierarchy construction is the number of LLM calls. Let $C$ be the number of contributions (\S\ref{subsec:representing:papers}), $b$ the branching factor, and $L = O(\log_b C)$ the maximum depth for a near-balanced tree (\S\ref{subsec:depth}).
\nameAlg{} requires $O\left(C/b\right)$ LLM calls for both top-down and bottom-up variants. 
The table here also includes LLM-intensive baselines discussed in \S\ref{subsec:pure-llm:baselines}, \baselineOne{} makes $O(C / l)$ calls (with $l$ as batch size), offering lower complexity but at the cost of reduced quality. In contrast, \baselineTwo{} achieves higher accuracy but requires $O(C \log_b C)$ LLM calls due to root-to-leaf traversals during insertion.
These theoretical results are validated empirically (\S\ref{subsec:pure-llm:baselines}): \baselineTwo{} makes $\times 190$ more calls compared to \nameAlg{}.

\newcommand{\dataOne}{\textbf{SciPile}}
\newcommand{\dataTwo}{\textbf{SciPileLarge}}

\section{Experimental Setup and Results}

We describe our experimental setup, including the diverse paper collection used for our experiments (\S\ref{subsec:collecting:papers}) and the evaluation framework (\S\ref{sec:eval:details}).


\subsection{Collection of Science Papers}
\label{subsec:collecting:papers}

We compile a collection of scientific papers spanning domains such as computer science, neuroscience, biology, oceanography, and their interdisciplinary intersections.
Our initial analysis focuses on a smaller set of approximately $2$K papers (referred to as \dataOne), allowing for rapid iteration over design choices and assessment of scalability. We then extend our analysis to a larger collection of $10K$ papers, referred to as \dataTwo. Details on data collection and filtering are provided in \S\ref{appendix:collecting:papers}.

\begin{wraptable}[14]{r}{0.49\textwidth}
\vspace{-0.33cm}
\centering
\footnotesize
\setlength{\tabcolsep}{2pt}
\begin{tabular}{@{}lcc@{}}
\toprule
Primary domain & \dataOne{} & \dataTwo{}  \\
\midrule
Medicine              & 29.6 & 24.2 \\
Computer Science      & 21.5 & 23.8 \\
Physics               & 22.1 & 16.0 \\
Materials Science     &  7.9 & 11.5 \\
Environmental Science &  5.4 &  3.9 \\
Other domains         & 13.5 & 20.6 \\
\bottomrule
\end{tabular}
\renewcommand{\arraystretch}{1}
\caption{Distribution (\%) of Semantic Scholar primary-domain labels in \dataOne{} and \dataTwo{}. The five most frequent domains are shown separately; \textit{Other domains} aggregates all remaining labels.}
\label{tab:corpus-domain-distribution}
\end{wraptable}
\noindent
\textbf{Disciplinary distribution.}
Table~\ref{tab:corpus-domain-distribution} reports the five most frequent Semantic Scholar primary-domain labels in each corpus. We design the collection to make it  multidisciplinary: no single domain accounts for more than 30\% of either collection. Their mixtures also differ; for example, \dataTwo{} contains larger proportions of Computer Science and Materials Science papers, whereas \dataOne{} contains larger proportions of Medicine and Physics papers. These coarse labels characterize corpus composition and need not map one-to-one to the finer-grained subfields induced by \nameAlg{}.



\definecolor{lightblue}{RGB}{235,244,250}
\definecolor{lightgreen}{RGB}{237,246,240}
\definecolor{lightpurple}{RGB}{242,240,247}
\definecolor{highlight}{RGB}{192,0,0} 

\definecolor{highlight_green}{RGB}{0,192,0}
\newcommand{\specialcell}[2][c]{
\begin{tabular}[#1]{@{}c@{}}#2\end{tabular}
}

\newcommand{\arrowdownright}{\hspace{0.03cm}\reflectbox{\rotatebox[origin=c]{180}{$\Rsh$}}\hspace{0.03cm}}

\newcommand{\smallfont}[1]{{\fontsize{6.5}{5}\selectfont #1}}

\subsection{Evaluation as Utilization}
\label{subsec:evaluation:utiliization}
\label{sec:eval:details}
\vspace{-0.2cm}
Ideally, hierarchy quality would be evaluated against a gold standard---but no such reference exists, and scientific literature continually evolves. As a result, we adopt an evaluation framework based on \textit{utilization}, independent of fixed ground truth.

We assess hierarchy quality by measuring \textit{how well it supports navigation and content discovery}. Specifically, we use an LLM-based agent to locate target papers via tree traversal, tracking accuracy at each level and across the full hierarchy. A stronger hierarchy should better capture conceptual relationships and improve information-seeking efficiency. While our evaluation focuses on retrieval, the hierarchy’s utility extends beyond that.

Our evaluation design involves two choices: (a) queries and (b) an evaluation model.
For (a), we sample paper titles and abstracts. Although we considered generating language questions from papers, pilot studies showed both approaches yield similar results, so we use the simpler method. 
For (b), we use \texttt{Qwen2.5-32b-instruct}, the open model with the highest GPT-4 agreement once our own summarizer, Llama-3.3-70B, is excluded to avoid self-preference (\S\ref{appendix:evaluation-pilot}).



The process starts at the root: given a query and cluster descriptions (Fig.~\ref{fig:prompt-evaluation}), the LLM selects the most relevant cluster. If it contains the target paper, traversal continues recursively through subclusters until the correct paper-level node is reached.
We report two metrics: \textbf{Strict-Acc}, the fraction of cases where the model finds the target node, and \textbf{L1-Acc}, which measures how often it correctly identifies the top-level subtree containing the target. Unless noted otherwise, we report the mean $\pm$ one standard deviation over five independent evaluation runs.

Because branching factor jointly controls local choice complexity and hierarchy depth, we treat it as an aspect of hierarchy quality rather than normalizing it away. We therefore report branching factors explicitly and complement navigation accuracy with the branching-factor-independent manual relation evaluation in \S\ref{subsection:additional_ana}.

\vspace{-0.2cm}
\section{Empirical Findings and Results}
\label{sec:findings}
\label{sec:hyperparams}
\vspace{-0.2cm}
We conduct a series of experiments to evaluate \nameAlg{} (\S\ref{sec:muhan:approach}) in various settings. 
The hyperparameters for \nameAlg{} are detailed in \S\ref{appendix:hyperparams}. In \S\ref{sec:visualization} we show a slice of the final hierarchy generated by \nameAlg{} on the \dataTwo{}.


\subsection{Comparison to Existing Taxonomies}
\label{sec:compare-alg}
We compare \nameAlg{} against two representative external baselines covering both existing taxonomies and algorithms: (1)~OpenAlex~\citep{priem2022openalex}, a widely adopted four-level academic taxonomy, where we retrieve its existing labels for our datasets to construct the OpenAlex hierarchy; and (2)~TaxoAdapt~\citep{kargupta-etal-2025-taxoadapt}, an LLM-based method that iteratively expands taxonomies into DAG-structured hierarchies. Among prior algorithms discussed in \S\ref{sec:related:work}, TaxoAdapt is the most directly comparable, as it operates at the document level and scales to corpora of thousands of papers.

Table~\ref{tab:external-baselines} presents the comparison on \dataTwo{}. Against OpenAlex (Table~\ref{tab:external-baselines}, left), \nameAlg{} attains higher mean Strict-Acc for all three contribution types; against TaxoAdapt (Table~\ref{tab:external-baselines}, right), it attains higher mean accuracy on both aligned dimensions. The baselines also differ notably from \nameAlg{} in \textbf{branching factor}: our hierarchies have a maximum branching factor of at most 30, whereas OpenAlex and TaxoAdapt exceed this by more than ten times. However, this substantially greater breadth does not translate into higher Strict-Acc. \emph{Overall, \nameAlg{} achieves higher mean accuracy than both external baselines while producing considerably more compact hierarchies.}

\vspace{-0.2cm}
\begin{table*}[ht]
\centering
\small
\begin{minipage}[c]{0.5\textwidth}
\centering
\setlength{\tabcolsep}{2pt}
\renewcommand{\arraystretch}{1.3}
\begin{tabular*}{\textwidth}{@{\extracolsep{\fill}}lccc@{}}
\toprule
\multirow{2}{*}{Method} & \multicolumn{2}{c}{Accuracy (\%)} & \multirow{2}{*}{\makecell{Max.\\BF}} \\
\cmidrule(lr){2-3}
& Strict-Acc & L1-Acc & \\
\midrule
\nameAlg{} (Problem) & \textbf{43.7} \smallfont{$\pm$ 6.5} & \textbf{85.8} \smallfont{$\pm$ 4.2} & 26 \\
\nameAlg{} (Solution) & 24.7 \smallfont{$\pm$ 4.8} & 65.8 \smallfont{$\pm$ 2.5} & 28 \\
\nameAlg{} (Result) & 27.6 \smallfont{$\pm$ 4.6} & 69.8 \smallfont{$\pm$ 2.1} & 30 \\
\midrule
OpenAlex & 20.4 \smallfont{$\pm$ 3.2} & 78.8 \smallfont{$\pm$ 7.0} & 313 \\
\bottomrule
\end{tabular*}
\end{minipage}
\hfill
\begin{minipage}[c]{0.45\textwidth}
\centering
\setlength{\tabcolsep}{2pt}
\renewcommand{\arraystretch}{0.9}
\begin{tabular*}{\textwidth}{@{\extracolsep{\fill}}lccc@{}}
\toprule
\multirow{2}{*}{Method} & \multicolumn{2}{c}{Accuracy (\%)} & \multirow{2}{*}{\makecell{Max.\\BF}} \\
\cmidrule(lr){2-3}
& Strict-Acc & L1-Acc & \\
\midrule
\rowcolor{lightpurple} \multicolumn{4}{@{}l}{\textit{Contribution type: Problem}} \\
\noalign{\vskip 2pt}
\nameAlg{} & \textbf{43.7} \smallfont{$\pm$ 6.5} & 85.8 \smallfont{$\pm$ 4.2} & 26 \\
TaxoAdapt & 37.9 \smallfont{$\pm$ 3.9} & 82.5 \smallfont{$\pm$ 2.6} & 1405 \\
\midrule
\rowcolor{lightgreen} \multicolumn{4}{@{}l}{\textit{Contribution type: Solution}} \\
\noalign{\vskip 2pt}
\nameAlg{} & \textbf{24.7} \smallfont{$\pm$ 4.8} & 65.8 \smallfont{$\pm$ 2.5} & 28 \\
TaxoAdapt & 14.3 \smallfont{$\pm$ 9.5} & 41.5 \smallfont{$\pm$ 4.6} & 732 \\
\bottomrule
\end{tabular*}
\end{minipage}
\caption{Comparison of \nameAlg{} with existing methods on \dataTwo. BF denotes Branching Factor. \textit{Left}: OpenAlex provides a single contribution-agnostic taxonomy with a significantly higher branching factor. \textit{Right}: TaxoAdapt's \textit{task} and \textit{methodology} dimensions are aligned with our Problem and Solution attributes respectively. \nameAlg{} outperforms both baselines. A detailed case study in \S\ref{appendix:openalex} further illustrates OpenAlex's limitations.}
\label{tab:external-baselines}
\end{table*}

\begin{wraptable}[20]{r}{0.5\textwidth}
\setlength{\tabcolsep}{2pt}
\centering
\footnotesize
\vspace{-1cm}
\definecolor{lightblue}{RGB}{235,244,250}
\definecolor{lightgreen}{RGB}{237,246,240}
\definecolor{lightpurple}{RGB}{242,240,247}
\renewcommand{\arraystretch}{0.85}
\begin{tabular}{@{}l*{3}{c}@{}}
\toprule
\multirow{2}{*}{Method} & \multicolumn{2}{c}{Accuracy (\%)} & {LLM Cost} \\
\cmidrule(lr){2-3} \cmidrule(lr){4-4}
& Strict-Acc $\uparrow$ & L1-Acc $\uparrow$ & \specialcell{\# of Input\\Tokens $\downarrow$} \\
\midrule
\rowcolor{lightpurple} \multicolumn{4}{@{}l}{\textit{Contributions type: Problem Statement}} \\
\midrule
\nameAlg{} & {\textbf{43.7} \smallfont{$\pm$ 6.5}} & 85.8 \smallfont{$\pm$ 4.2} & 7451 \\
\arrowdownright Top-down & 41.5 \smallfont{$\pm$ 8.2} & {\textbf{86.5} \smallfont{$\pm$ 5.6}} & 8990 \\
\arrowdownright Bottom-up & 26.2 \smallfont{$\pm$ 5.4} & 41.9 \smallfont{$\pm$ 4.0} & 5924 \\
\midrule
\rowcolor{lightgreen} \multicolumn{4}{@{}l}{\textit{Contributions type: Solution Statement}} \\
\midrule
\nameAlg{} & {\textbf{24.7} \smallfont{$\pm$ 4.8}} & {\textbf{65.8} \smallfont{$\pm$ 2.5}} & 7653 \\
\arrowdownright Top-down & 22.4 \smallfont{$\pm$ 3.5} & 52.3 \smallfont{$\pm$ 3.0} & 4032 \\
\arrowdownright Bottom-up & 23.9 \smallfont{$\pm$ 3.3} & 51.3 \smallfont{$\pm$ 3.1} & 6150 \\
\midrule
\rowcolor{lightblue} \multicolumn{4}{@{}l}{\textit{Contributions type: Results Statement}} \\
\midrule
\nameAlg{} & {\textbf{27.6} \smallfont{$\pm$ 4.6}} & {\textbf{69.8} \smallfont{$\pm$ 2.1}} & 6457 \\
\arrowdownright Top-down & 19.7 \smallfont{$\pm$ 4.0} & 54.0 \smallfont{$\pm$ 3.3} & 5380 \\
\arrowdownright Bottom-up & 23.6 \smallfont{$\pm$ 2.7} & 55.2 \smallfont{$\pm$ 2.9} & 4731 \\
\bottomrule
\end{tabular}
\renewcommand{\arraystretch}{1}
\caption{
    Evaluation of \nameAlg{} and baselines on the 10K (\dataTwo{}) dataset. \nameAlg{} achieves high accuracy across contribution types, validating our hybrid design.
}
\label{tab:exp-results-large:main:text}
\end{wraptable}
\vspace{-0.3cm}
\subsection{Ablation Against Special Cases}
\label{sec:baseline:topdown:bottomup}

We study two special cases of \nameAlg{}: a purely top-down strategy and a purely bottom-up approach. These ablations isolate the strengths and limitations of each component.

As shown in Table~\ref{tab:exp-results-large:main:text}, \nameAlg{} attains the highest mean Strict-Acc for all three contribution types, though its margin over the top-down ablation lies within overlapping standard deviations. The decisive differences are elsewhere: top-down trails by over 13 L1-Acc points on \textit{solution} and \textit{result} contributions, and bottom-up collapses on \textit{problem} contributions (41.9 vs.\ 85.8 L1-Acc), leaving \nameAlg{} as the only variant that is strong across all settings.
These improvements come at input-token counts within a factor of two of the ablations, and in fact lower than top-down for problem contributions.
\emph{Together, these results show that combining top-down and bottom-up signals yields the most consistently strong hierarchies across contribution types.}

\vspace{-0.1em}
\subsection{Comparison to LLM-intensive Baselines}
\label{subsec:pure-llm:baselines}

We compare \nameAlg{} against two LLM-intensive baselines motivated by the hypothesis that LLMs can make high-quality local placement decisions that collectively yield a coherent global hierarchy. The key tradeoff of this approach is cost, as it requires \textit{many} LLM calls. We refer to these baselines as \baseline{} (pronounced ``flimsy'') and describe two variants below.
For both variants, we use \texttt{gpt-4o} to extract contributions (\S\ref{subsec:representing:papers}) and \texttt{Llama-3.3-70B-Instruct} to place them into the hierarchy.

Both methods begin from a shared seed hierarchy, initialized using the Wikipedia hierarchy of sciences\footnote{\href{https://en.wikipedia.org/wiki/Branches_of_science}{en.wikipedia.org/wiki/Branches\_of\_science}} and refined through several manual adjustments detailed in \S\ref{appendix:seed:hierarchy}.
\baselineOne{}  expands the seed hierarchy by aggregating all unique contributions into batches of 100 and inserting each batch in parallel via separate threads, where the LLM inserts contributions into a cloned copy of the seed hierarchy; the resulting hierarchies are then merged via a Python script without additional LLM calls. \baselineTwo{}  constructs the hierarchy by inserting one contribution at a time through layer-by-layer prompting, where the model traverses the tree and selects one of four actions: \textit{Go down}, \textit{Add sibling}, \textit{Make parent}, or \textit{Discard} (Fig.~\ref{fig:incremental:actions}). To prevent overly specific nodes from being placed too high, node-creation actions are disabled above layer~3. Pilot studies revealed that replacing top-level labels with descriptive definitions (Fig.~\ref{fig:descriptive-definitions}) significantly improved placement accuracy.

\begin{wraptable}{r}{0.5\textwidth}
  \vspace{-0.3cm}
  \setlength{\tabcolsep}{2pt}
  \renewcommand{\arraystretch}{0.85}
  \centering
  \footnotesize
  \begin{tabular}{@{}lccc@{}}
    \toprule
    Method & Strict-Acc & L1-Acc & \# of Calls $\downarrow$ \\
    \midrule
    \rowcolor{lightblue} \multicolumn{4}{@{}l}{\textit{Topic contributions}} \\
    \midrule
    \nameAlg{} & {\textbf{14.9} \smallfont{$\pm$ 2.7}} & {\textbf{65.7} \smallfont{$\pm$ 4.4}} & 322 \\
    \arrowdownright Top-down & 14.5 \smallfont{$\pm$ 4.7} & 62.5 \smallfont{$\pm$ 7.4} & 322 \\
    \arrowdownright Bottom-up & 13.9 \smallfont{$\pm$ 5.3} & 54.4 \smallfont{$\pm$ 12.7} & 322 \\
    \graymidrule
    \arrowdownright \baseline{} (par) & 4.0 \smallfont{$\pm$ 2.8} & 32.0 \smallfont{$\pm$ 6.3} & 226 \\
    \arrowdownright \baseline{} (inc) & {\textbf{18.0 \smallfont{$\pm$ 5.3}}} & {\textbf{91.0 \smallfont{$\pm$ 4.0}}} & {\color{highlight}{61K}} \\
    \bottomrule
  \end{tabular}
  \caption{
  Results for \nameAlg{}, \baseline{} (\textbf{par}allel) and \baseline (\textbf{inc}remental) using \textit{Topic} as contribution type.
  While \baseline{} (inc) achieves the highest accuracy, it requires $\approx$190$\times$ more LLM calls. Full results in \S\ref{appendix:topic_result}.
  }
  \label{tab:topics-2k}
\end{wraptable}
\noindent
\textbf{Results: LLM-only approaches are expensive and less scalable.}
Table~\ref{tab:topics-2k} reports results on the simplest contribution type (``topic'') in \dataOne{}, chosen due to the substantial computational cost of the LLM-only baselines.
While \baselineTwo{} slightly outperforms \nameAlg{} in strict accuracy, it does so at the cost of a \textit{massive} increase in LLM calls, rendering it impractical at scale. Consequently, despite strong accuracy, \baselineTwo{} does not scale to larger corpora.
\emph{In contrast, \nameAlg{} achieves competitive performance while scaling efficiently to substantially larger paper collections.}
\vspace{-0.3cm}
\subsection{Additional Analyses}
\label{subsection:additional_ana}
We briefly summarize additional analyses, with further details for some experiments provided in the appendix.

\begin{wraptable}[12]{r}{0.48\textwidth}
\vspace{-0.45cm}
\centering
\footnotesize
\setlength{\tabcolsep}{4pt}
\renewcommand{\arraystretch}{0.9}
\begin{tabular}{@{}lcc@{}}
\toprule
Corpus & Strict-Acc (\%) & L1-Acc (\%) \\
\midrule
STEM           & 52.4 & 84.9 \\
Biomedical     & 47.8 & 86.7 \\
Social Science & 59.5 & 92.5 \\
\bottomrule
\end{tabular}
\caption{
Navigation accuracy of \nameAlg{} on three independently collected,
approximately $10$K-paper domain-focused corpora, using
\textit{problem} contributions. Results are from a single evaluation run.
}
\label{tab:domain-robustness}
\end{wraptable}
\noindent
\textbf{Robustness across scientific domains.}
In addition to constructing a unified hierarchy over a broad scientific corpus,
we evaluate whether \nameAlg{} supports domain-focused organization.
We construct three independent corpora, each containing approximately $10$K
papers and respectively covering STEM, biomedical science, and social science.
Table~\ref{tab:domain-robustness} reports results for hierarchies built from
\textit{problem} contributions.
\nameAlg{} achieves consistently strong navigation accuracy across all three
settings. These results suggest that domain-focused hierarchies can complement
the unified hierarchy when users seek finer-grained organization within a
known field.

\noindent
\textbf{Performance is stable across cluster schedules.}
As discussed in \S\ref{subsec:depth}, we set the hierarchy depth according to
corpus scale. To assess sensitivity to the layer-wise cluster schedule, we fix
the hierarchy depth at $L=4$ and the top level at six clusters while varying
the lower-level schedule from coarse to fine.
Table~\ref{tab:cluster-schedule-sensitivity} shows that increasing the number
of bottom-level clusters from 625 to 2,500 results in only a marginal change
in Strict-Acc, indicating limited sensitivity to this design choice.





\noindent
\textbf{Detailed prompts significantly improve hierarchy quality. }To demonstrate this, we compare two prompt types. The first is a ``detailed'' prompt—carefully curated with comprehensive instructions and reminders, which we use for all main experiments in this paper. The second is a ``simplified'' prompt containing only the core task description. The detailed prompt substantially outperforms the simplified version across all scenarios (\S\ref{appendix:prompt}).

\begin{wraptable}[12]{r}{0.4\textwidth}
\vspace{-0.3cm}
\centering
\footnotesize
\setlength{\tabcolsep}{4pt}
\renewcommand{\arraystretch}{0.9}
\begin{tabular}{@{}lc@{}}
\toprule
Cluster schedule & Strict-Acc (\%) \\
\midrule
$(6,30,150,625)$  & 45.2 \\
$(6,40,276,1250)$ & 43.7 \\
$(6,60,500,2500)$ & 43.5 \\
\bottomrule
\end{tabular}
\caption{Sensitivity to layer-wise cluster schedules on the 10K-paper corpus using \textit{problem} contributions. Results are from a single evaluation run.}
\label{tab:cluster-schedule-sensitivity}
\end{wraptable}

\noindent
\textbf{Manual relation evaluation provides non-LLM evidence of hierarchy quality.}
To complement the navigation-based evaluation, we manually assess 300
parent--child relations sampled from three independently generated
\dataTwo{} hierarchies. We sample 100 relations per hierarchy, stratified
evenly across its four depths (25 per depth). Of the 300 relations, 229 are strictly hierarchical, 43 are semantically related but
not strictly hierarchical, and 28 are incorrect or incoherent. Thus,
90.7\% of the sampled relations are at least meaningfully related. Unlike navigation accuracy, this assessment judges each relation directly, without a choice among siblings, and is thus less sensitive to branching factor.

\noindent
\textbf{Embedding quality varies a lot across models.} We evaluate three models for \embedder: Qwen’s gte-Qwen2-7B-instruct~\citep{li2023towards}, OpenAI’s text-embedding-3-large, and text-embedding-ada-002. The first two perform similarly, whereas text-embedding-ada-002 produces markedly weaker results. We select gte-Qwen2-7B-instruct for its strong performance and open weights (results in \S\ref{appendix:embedding}).

\noindent
\textbf{Human and curated-hierarchy checks support evaluator reliability.}
As our navigation accuracy relies on an LLM evaluator, we conduct two complementary validation studies. (1) In a manual analysis of 200 navigation errors, only 9 were clear evaluator mistakes; in 39 cases, both the evaluator's choice and the original hierarchy path were reasonable; and in the remaining 152 cases, the human annotator agreed with the evaluator. (2) As an external calibration, we apply the same evaluation protocol to a human-curated, two-level \href{https://orkg.org/}{ORKG} hierarchy containing 4.4K engineering papers. Given each paper's title and abstract as the query, the evaluator reaches the corresponding paper through the hierarchy in 83\% of cases.

\section{Conclusion}
\paragraph{Future applications:}
Our work opens several promising directions for future research. One key opportunity is to use the constructed hierarchies as tools for exploratory analysis across scientific domains. They can aid academic institutions and funding bodies in identifying emerging trends and underexplored areas, and can be adapted for domain-specific analyses that capture the unique structure of individual fields. This approach not only deepens our understanding of scientific progress but also provides a new lens for organizing the vast and growing body of scholarly work.

\noindent
\textbf{Conclusions:}
We introduced a scalable framework for organizing scientific literature into
multi-level hierarchies that connect broad research domains with fine-grained
contributions and individual papers. Our method, \nameAlg{}, combines
embedding-based clustering with LLM summarization in a hybrid top-down and
bottom-up design, balancing semantic quality with scalability. To reflect the
multifaceted nature of scientific work, we construct separate hierarchies for
different contribution types, including problems, solutions, and results.
Experiments on unified and domain-focused corpora demonstrate strong
navigation performance across scientific domains, while manual relation assessment provides complementary non-LLM evidence of hierarchy quality. By providing a bird's-eye view of how research activity is organized, these hierarchies complement conventional query-based retrieval and support exploratory analysis of the scientific landscape.





\section*{Ethics Statement}
In our work, all data and models are accessed via licenses that grant us free and open access for research purposes. 
Expert annotations are provided by the paper's authors, who have contributed their efforts without compensation. 
We have not observed any harmful content in either the scholarly papers or the content generated by LLMs. 
On the other hand, since our resulting hierarchy reflects the distribution of scientific efforts across various fields, it offers a detailed map of where research activity is concentrated and where it is lacking. 
This nuanced view can guide decision-makers—such as government agencies and academic institutions—in making more informed choices about resource allocation. 
By highlighting underexplored yet promising areas alongside well-established fields, the hierarchy helps ensure that funding, support, and strategic initiatives are distributed more equitably. 
Ultimately, this balanced approach can foster innovation and drive progress in areas that might otherwise be overlooked, leading to a more inclusive and socially beneficial advancement of science.

\section*{Acknowledgments}
This work is supported by ONR grant (N0001424-1-2089) and
Defense Advance Research Projects Agency (DARPA) under Contract No. HR001125C0304.
Any opinions, findings and conclusions or recommendations expressed in this material are those of the author(s) and do not necessarily reflect the views of DARPA.
We sincerely thank Jiefu Ou and the broader JHU CLSP
community for discussions and inspiration.

\bibliographystyle{colm2026_conference}
\bibliography{ref_custom,ref}

\newpage

\appendix
\onecolumn

\section{Limitations}
\label{appendix:limitations}
Although we evaluated our pipeline on $10K$ papers, this is still far from the true scale of scientific literature. We hope future work will enhance our approach to handle more realistic scales. Additionally, while our evaluation framework shows potential for efficient information discovery, it may have its own weaknesses and biases. We partially address this through human verification, but these studies rely on a relatively small number of annotators. Larger-scale, more systematic human assessment would further strengthen confidence in the quality and reliability of the inferred hierarchies.

\section{Additional Related Work}
\label{appendix:relate_work}

We include additional related work here because of the space limitation in the main text. 

\noindent
\textbf{Clustering with LLMs:}
Recent advances in clustering methodologies augmented by LLMs have demonstrated effective ways to generate interpretable groupings of text. 
For example, \cite{viswanathan2024large,katz2024knowledge} apply few-shot clustering and thematic grouping to partition scientific literature into meaningful subtopics, while \cite{zhang2023clusterllm,wang2023goal} further refine these techniques by aligning clustering outcomes with natural language explanations and user intent. 
Other recent work iteratively refines cluster representations by replacing cluster centroids or summary points with LLM-generated natural language descriptions and inclusion criteria, thereby inducing more abstract, interpretable concepts over multiple clustering rounds \citep{DBLP:conf/chi/LamTLHB24,diaz2025k}.
While these approaches improve clustering quality by using LLMs at various stages, they mostly result in flat groupings rather than hierarchical structures. Our approach builds on this by using LLMs to cluster documents and organizing these clusters into a structured hierarchy. 

\paragraph{Structured knowledge representation:}
Understanding and organizing knowledge is a fundamental pursuit in both artificial and human intelligence~\citep{dahlberg1993knowledge}.
Abstraction hierarchies, such as WordNet for lexical semantics~\citep{miller1995wordnet}, ConceptNet for commonsense reasoning~\citep{DBLP:conf/aaai/SpeerCH17}, and Probase for large-scale concept representation~\citep{DBLP:conf/sigmod/WuLWZ12}, have proven to be powerful tools for structuring information.
Similarly, modern tabular reasoning leverages structured representations to facilitate systematic inference and knowledge retrieval, demonstrating that such structure remains crucial~\citep{DBLP:conf/iclr/0002ZLEP0MFSLP24}.

\paragraph{Comparison with existing hierarchical systems:} 
\href{https://www.scopus.com/home.uri}{Scopus} uses a \textbf{fixed-depth} (3-layer) hierarchy based on research field names (ASJC Codes). Importantly, these codes are assigned at the journal level rather than to individual papers. Papers inherit classifications from their publishing journals, meaning the hierarchy is not derived from the actual paper content. \href{https://pubmed.ncbi.nlm.nih.gov/}{PubMed MeSH terms} provides hierarchical labeling for PubMed publications, but it functions at the level of \textbf{keywords (few tokens)} rather than leveraging the full richness of natural language from science papers. Crucially, it is organized around a fixed set of controlled terms rather than the actual semantic content of the papers, limiting its suitability for constructing dynamic or corpus-specific hierarchies. Additionally, because MeSH is manually curated, it introduces indexing delays—papers are only labeled after publication—and is subject to human bias, as noted by ~\cite{hadfield2020delay}.
\href{https://www.microsoft.com/en-us/research/project/microsoft-academic-graph/}{Microsoft Academic Graph (MAG)}, though \textbf{discontinued} in 2021, offered a rich graph-based structure connecting papers and authors. Its hierarchical classification derived primarily from citation patterns and machine learning clustering rather than semantic paper content, which limited cluster interpretability.

\clearpage

\section{Evaluation Framework}
\label{appendix:evaluation}

\begin{figure}[htbp]
    \centering
    \includegraphics[width=1\linewidth,trim=0cm 0.9cm 0cm 0cm]{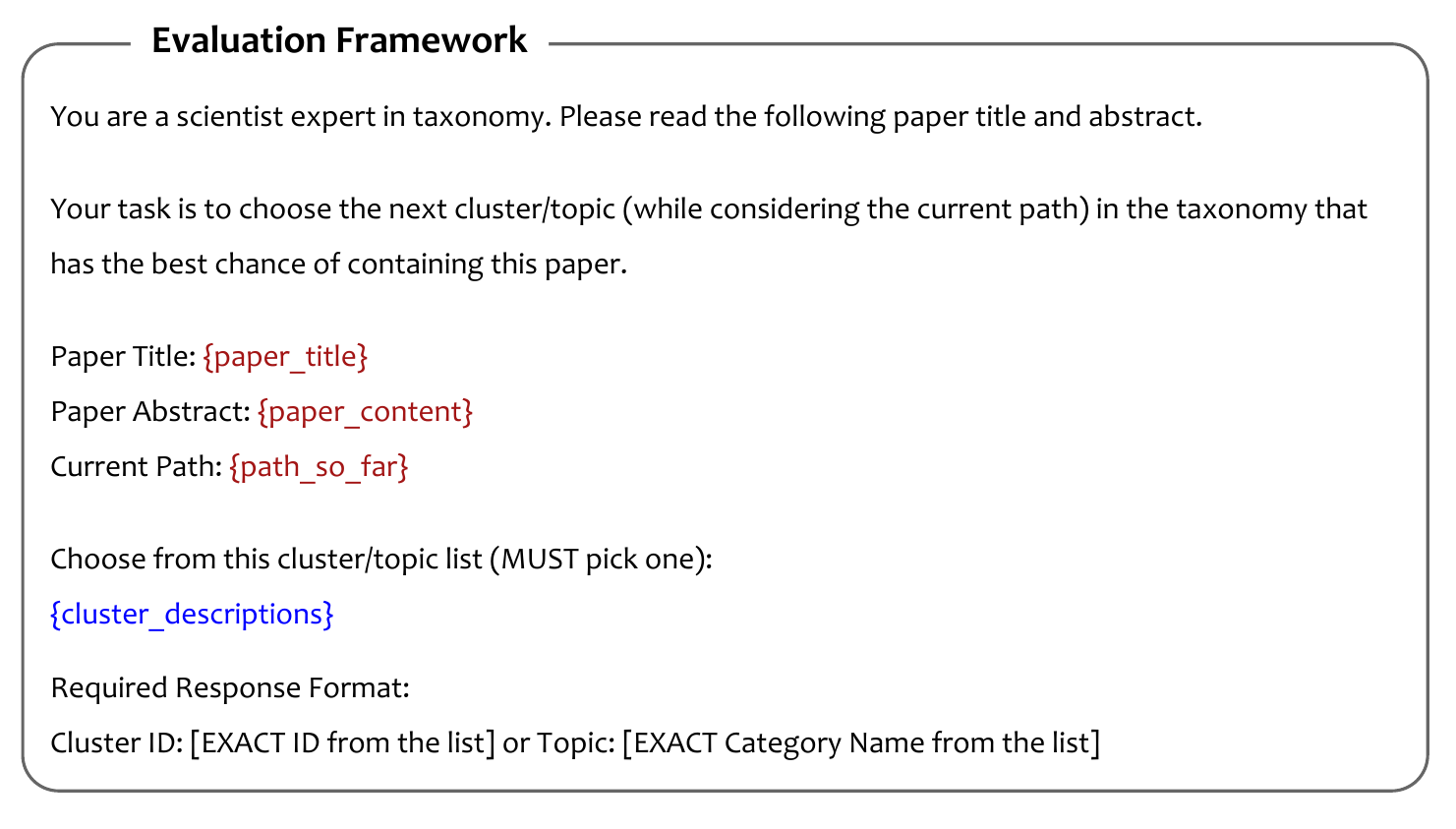}
    \caption{Prompt used for Evaluation}
    \label{fig:prompt-evaluation}
\end{figure}

We provide more context on our evaluation. 
As discussed in \S\ref{sec:eval:details}, we use randomly-sampled papers (title/abstract) as a query. The evaluator LLM goes through the hierarchy, starting from the root node and iteratively selects the relevant nodes to traverse. The prompt for each decision is shown in Fig.\ref{fig:prompt-evaluation}. 

\subsection{Pilot Experiment for Evaluator Choice}
\label{appendix:evaluation-pilot}
One question is, \textbf{which LLM should we use for evaluation?}
As discussed in \S\ref{subsec:evaluation:utiliization}, we chose \texttt{Qwen2.5-32b-instruct} for its strong instruction-following capabilities. In pilot experiments, Qwen showed a high consistency against GPT4 score, compared to other open-weight models. Here's a summary of that experiment: We evaluated one of the hierarchies produced by \nameAlg{} using different models, including GPT-4. Assuming GPT-4 has the highest accuracy, we sought alternative models with the greatest consistency against it, as frequent evaluations with GPT4 are costly. Table~\ref{tab:consistency:gpt4} presents the results. As it can be observed, Llama has the highest agreement, but we suspect bias since the hierarchy was also constructed with Llama. To avoid this, we selected the next best model, Qwen2.5-32b-instruct, for evaluation. 


\begin{table}[h!]
\centering
\small
\begin{tabular}{lc}
\toprule
\textbf{Evaluator LLM} & \textbf{Agreement with GPT4} \\
\midrule
GPT-3.5 & 39.6 \\
GPT4-mini & 59.2 \\
Gemma3-24b-it & 62.1 \\
Qwen2.5-32b-instruct & 66.5 \\
Llama 3.3 70B & 72.4 \\
\bottomrule
\end{tabular}
\caption{Agreement of different evaluator LLMs against GPT4.}
\label{tab:consistency:gpt4}
\end{table}

\subsection{Validation of LLM-based Evaluation}
\label{appendix:evaluation-val}

As we discuss in \S\ref{subsec:evaluation:utiliization}, to validate our evaluation framework, a Computer Science PhD student analyzes 200 error cases (50 cases per layer). For each case, the annotator determines whether the error comes from the LLM evaluator or from the hierarchy itself. The analysis reveals three types of cases. First, only 9 cases (4.5\%) are clear evaluator errors. Second, in 39 cases (19.5\%), both the evaluator's choice and the hierarchy path are reasonable, which is expected for interdisciplinary works. Third, in the remaining 152 cases (76\%), the evaluator agrees with the human annotator. These results confirm the reliability of our LLM-based evaluation approach.

To further validate our LLM-based evaluation approach, we downloaded the annotations from the \href{https://orkg.org/}{Open Research Knowledge Graph} (ORKG). On this website, papers are curated entirely by volunteers who are strongly familiar with the topics of the papers. We use the subset of the ORKG data focused on the Engineering domain. This led to a collection of 4.4K papers that are organized in a 2-layer hierarchy. Treating this data as a high-quality hierarchy, the question is whether our evaluation would assign it a high score.
We ran our evaluation experiment with \texttt{Qwen2.5-32B-Instruct} as the LLM-as-a-judge. Similar to our setup from the paper, we use paper title/abstract as queries, and require the evaluator to traverse the hierarchy by incrementally making the most appropriate choice between all possible cluster candidates. 
Our results show that the evaluator model has an accuracy of 83\% (i.e., in 83\% of the runs it identified the correct paper). This indicates that our evaluation metric is able to assign a high score to a good hierarchy.

\subsection{Comparison with OpenAlex taxonomy}
\label{appendix:openalex}

To provide independent validation of our hierarchy's quality, we compare against OpenAlex~\citep{priem2022openalex}'s taxonomy. OpenAlex is a fully open catalog of scholarly works that succeeded Microsoft Academic Graph (MAG) after its discontinuation in 2021. It provides a 4-level hierarchical taxonomy which is widely adopted in researches.

For the same set of papers (\dataTwo{}, 10K papers), we organize them according to their assigned labels in OpenAlex. Since both hierarchies contain four levels, no additional structural adaptation is required. We apply the same LLM evaluator used in our main experiments to traverse each hierarchy and attempt to locate papers based on the taxonomy structure.

\begin{table*}[th!]
\centering
\small
\begin{tabular}{@{}p{4.8cm}p{3.3cm}p{4.8cm}@{}}
\toprule
\textbf{Paper Title} & \textbf{OpenAlex} & \textbf{\nameAlg{}} \\
\midrule
\multicolumn{3}{l}{\textit{Subfield-Level (L2) Comparison}} \\
\hdashline
\noalign{\vspace{0.3em}}
Differentially Fed Dual-Band Base Station Antenna (5G) 
& Aerospace Engineering
& Next-Generation Wireless Communications \\[4pt]
Abnormalities in Diffusional Kurtosis Metrics Related to Head Impact 
& Epidemiology 
& Neurological Disorders and Brain Injury \\[4pt]
New Sea Spray Generation Function for Spume Droplets 
& Earth-Surface Processes 
& Oceanic Mixing and Air-Sea Flux \\[4pt]
Vortex Flow and Cavitation in Diesel Engines 
& Electrical Engineering 
& Multiphase Flow and Combustion \\
\midrule
\multicolumn{3}{l}{\textit{Topic-Level (L3) Comparison}} \\
\hdashline
\noalign{\vspace{0.3em}}
Visual Transformers and CNNs for Disease Classification 
& COVID-19 Diagnosis 
& Deep Learning for Medical Image Analysis \\[4pt]
A Cyclone-Centered Perspective on Sea Ice Motion 
& Arctic Ice Dynamics 
& Atmospheric Dynamics and Polar Cyclones \\[4pt]
Performance of Chip-Scale Optical Frequency Comb Generators 
& Fiber Laser Tech. 
& Integrated Photonics for Communications \\
\bottomrule
\end{tabular}
\caption{Side-by-side comparison of OpenAlex and \nameAlg{} classifications at corresponding hierarchy levels.}
\label{tab:openalex_vs_scychic}
\end{table*}


\newpage

\section{Extracting Paper Contributions}
\label{appendix:contributions}
As we discuss in \S\ref{subsec:representing:papers}, below are prompts and examples for extracting different contributions (\textit{problem}, \textit{solution}, \textit{result} and \textit{topics}) from papers' titles and abstracts. We utilize the GPT-4o model (\texttt{gpt-4o-2024-08-06}) to generate all contribution extractions along with detailed rationales explaining the extraction decisions.

\subsection{Prompt for Extracting \textit{Problem/Solution/Result} Contributions}
\label{appendix:contributions:1}

We use the prompt below to extract contributions from the paper's title and abstract. After finishing the extraction, the three contributions will be saved into the original \texttt{json} file. Please see \S\ref{subsec:representing:papers} for more information.

\begin{figure}[ht]
    \centering
    \includegraphics[width=1\linewidth,trim=0cm 0.5cm 0cm 0cm]{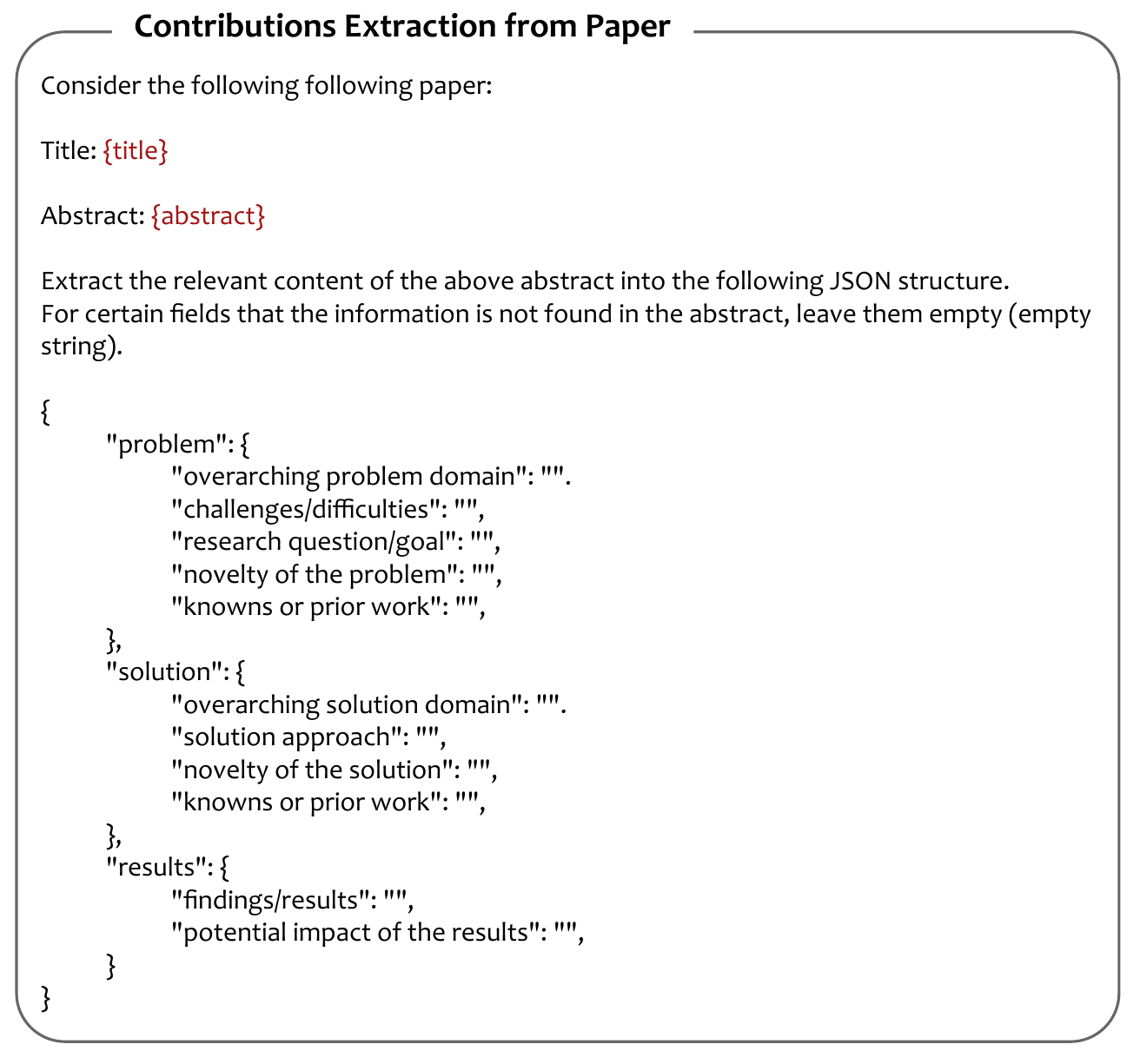}
    
    \caption{Prompt for extracting \textit{Problem/Solution/Result} contributions}
    \label{fig:prompt-abstract}
\end{figure}

\newpage
\subsection{Prompt for Extracting \textit{Topic} Contributions and Rationales}
\label{appendix:contributions:2}

This section has the prompt of generating topics and rationales from papers given their titles and abstracts. The prompt provides the model with a system role instruction that describes the task, title, and abstract, and also an example to get the specified output format.

\begin{figure}[htbp]
    \centering
    \includegraphics[width=1\linewidth,trim=0cm 0.5cm 0cm 0cm]{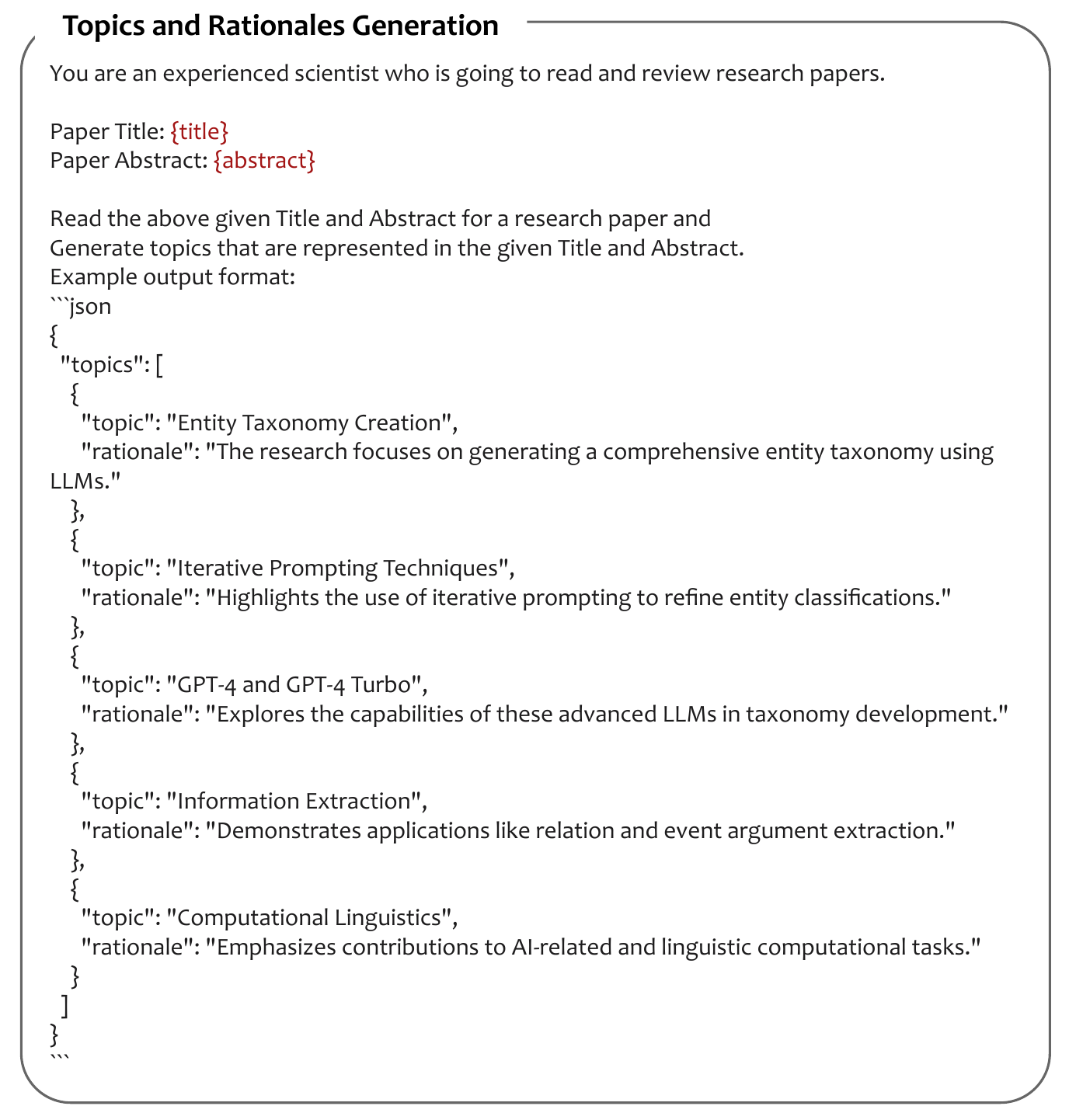}
    \caption{Prompt of \textit{Topic} and \textit{Rationale} Generation}
    \label{fig:prompt-rationale-generation}
\end{figure}

\subsection{Examples for \textit{Problem/Solution/Result/Topic} contributions extracted from papers}
Below we show examples of paper titles and abstracts, and different contributions (\textit{Problem/Solution/Result/Topic}) we extract by language model.

\lstdefinestyle{json}{
  basicstyle=\ttfamily\scriptsize,
  breaklines=true,
  columns=fullflexible,
  keepspaces=true,
  showstringspaces=false,
  tabsize=2,
  literate={\ \ }{{\ }}1,
  mathescape=true 
}

\definecolor{darkred}{rgb}{0.6,0,0}

\begin{table}[htbp]\setlength{\abovecaptionskip}{2pt}
    \small
    \centering
    \begin{tabular}{@{}p{13.9cm}@{}}
    \toprule
        \textit{\textbf{Problem/Solution/Result/Topic} contributions from scientific papers} \\
    \midrule
        \textcolor{brown}{\textbf{Title}}: \href{https://www.semanticscholar.org/paper/0b0f009cb4cee5da946bb6dfe0ae02127c337b1f}{Sixfold excitations in electrides} \\
        \textcolor{brown}{\textbf{Abstract}}: Due to the lack of full rotational symmetry in condensed matter physics, solids exhibit new excitations beyond Dirac and Weyl fermions, of which the sixfold excitations have attracted considerable interest owing to the presence of maximum degeneracy in bosonic systems. Here, we propose that a single linear dispersive sixfold excitation can be found in the electride Li$_{12}$ Mg$_3$ Si$_4$ and its derivatives. The sixfold excitation is formed by the floating bands of elementary band representation A@12a originating from the excess electrons centered at the vacancies (i.e., the 12a Wyckoff sites). There exists a unique topological bulk-surface-edge correspondence for the spinless sixfold excitation, resulting in trivial surface ``Fermi arcs'' but topological hinge arcs. All gapped k$_z$ slices belong to a two-dimensional higher-order topological insulating phase, which is protected by a combined symmetry T S$_{4z}$ and characterized by a quantized fractional corner charge Q$_{corner}$ = 3$|$e$|$/4. Consequently, the hinge arcs are obtained in the hinge spectra of the S$_{4z}$-symmetric rod structure. The state with a single sixfold excitation, stabilized by both nonsymmorphic crystalline symmetries and time-reversal symmetry, is located at the phase boundary and can be driven into various topologically distinct phases by explicit breaking of symmetries, making these electrides promising platforms for the systematic studies of different topological phases. \\
    \cdashline{1-1} \noalign{\vskip 0.2em}
    \end{tabular}
    
    \begin{tabular}{@{}p{4.45cm}@{\hspace{4pt}}p{4.45cm}@{\hspace{4pt}}p{4.45cm}@{}}
    \cellcolor{lightpurple}\textcolor{darkred}{\textbf{Contribution - Problem Statement}} & \cellcolor{lightgreen}\textcolor{darkred}{\textbf{Contribution - Solution Statement}} & \cellcolor{lightblue}\textcolor{darkred}{\textbf{Contribution - Result Statement}} \\
    \cellcolor{lightpurple}\begin{minipage}[t]{4.45cm}\vspace{-10pt}
\begin{lstlisting}[style=json]
{
  "overarching_problem_domain":
        "Condensed matter physics",
  "challenges/difficulties":
        "Lack of full rotational symmetry
        in solids leading to new excitations
        beyond Dirac and Weyl fermions",
  "research_question/goal":
        "Investigate sixfold excitations
        in electrides"
}
\end{lstlisting}
    \end{minipage} & 
    \cellcolor{lightgreen}\begin{minipage}[t]{4.45cm}\vspace{-10pt}
\begin{lstlisting}[style=json]
{
  "overarching_solution_domain":
        "Electrides and topological phases",
  "solution_approach":
        "Propose that a single linear
        dispersive sixfold excitation can
        be found in the electride
        Li$_{12}$Mg$_3$Si$_4$ and its derivatives",
  "novelty_of_the_solution":
        "Unique topological bulk-surface-edge 
        correspondence for the spinless
        sixfold excitation"
}
\end{lstlisting}
    \end{minipage} & 
    \cellcolor{lightblue}\begin{minipage}[t]{4.45cm}\vspace{-10pt}
\begin{lstlisting}[style=json]
{
  "findings/results":
        "The sixfold excitation is formed by
        floating bands of elementary band
        representation A@12a. All gapped
        k$_z$ slices belong to two-dimensional
        higher-order topological insulating
        phase, characterized by a quantized 
        fractional corner charge Q$_{corner}$ = 3$|$e$|$/4. 
        Hinge arcs are obtained in the hinge 
        spectra of the S$_{4z}$-symmetric rod
        structure.",
  "potential_impact_of_the_results":
        "Electrides are promising platforms
        for systematic studies of different
        topological phases."
}
\end{lstlisting}
    \end{minipage} \\
    \end{tabular}
    
    \begin{tabular}{@{}p{13.9cm}@{}}
    \cdashline{1-1} \noalign{\vskip 0.2em}
    \textcolor{darkred}{\textbf{Contribution - Topic}}: \textit{'Electrides', 'Electrides in Condensed Matter Physics', 'Higher-Order Topological Insulators', 'Nonsymmorphic Symmetries', 'Sixfold Excitation in Solids', 'Sixfold Excitations', 'Symmetry Breaking in Topological Materials', 'Topological Bulk-Surface-Edge Correspondence', 'Topological Phase Transitions', 'Topological Phases in Condensed Matter Physics', 'Topological Properties'} \\
    \bottomrule
    \end{tabular}
    \caption{Examples of extracted \textit{problem/solution/result/topic} contributions from scientific paper abstracts (example 1 of 2).}
    \label{tab:paper-contributions}
\end{table}

\begin{table}[htbp]\setlength{\abovecaptionskip}{2pt}
    \small
    \centering
    \begin{tabular}{@{}p{13.9cm}@{}}
    \toprule
        \textcolor{brown}{\textbf{Title}}: \href{https://www.semanticscholar.org/paper/0d61384ade721871d2183a9243346eb43dacf123}{The Tin Pest Problem as a Test of Density Functionals Using High-Throughput Calculations} \\
        \textcolor{brown}{\textbf{Abstract}}: At ambient pressure tin transforms from its ground-state semi-metal $\alpha$-Sn (diamond structure) phase to the compact metallic $\beta$-Sn phase at 13 $\bullet$ C (286K). There may be a further transition to the simple hexagonal $\gamma$-Sn above 450K. These relatively low transition temperatures are due to the small energy differences between the structures, $\approx$ 20 meV/atom between $\alpha$-and $\beta$-Sn. This makes tin an exceptionally sensitive test of the accuracy of density functionals and computational methods. Here we use the high-throughput Automatic-FLOW (AFLOW) method to study the energetics of tin in multiple structures using a variety of density functionals. We look at the successes and deficiencies of each functional. As no functional is completely satisfactory, we look Hubbard U corrections and show that the Coulomb interaction can be chosen to predict the correct phase transition temperature. We also discuss the necessity of testing high-throughput calculations for convergence for systems with small energy differences. \\
    \cdashline{1-1} \noalign{\vskip 0.2em}
    \end{tabular}
    
    \begin{tabular}{@{}p{4.45cm}@{\hspace{4pt}}p{4.45cm}@{\hspace{4pt}}p{4.45cm}@{}}
    \cellcolor{lightpurple}\textcolor{darkred}{\textbf{Contribution - Problem Statement}} & \cellcolor{lightgreen}\textcolor{darkred}{\textbf{Contribution - Solution Statement}} & \cellcolor{lightblue}\textcolor{darkred}{\textbf{Contribution - Result Statement}} \\ 
    \cellcolor{lightpurple}\begin{minipage}[t]{4.45cm}\vspace{-10pt}
\begin{lstlisting}[style=json]
{
  "overarching_problem_domain":
        "Density functionals and computational
        methods for phase transitions in
        materials.",
  "challenges/difficulties":
        "Small energy differences between
        phases of tin make it a sensitive
        test for the accuracy of density
        functionals.",
  "research_question/goal":
        "To study the energetics of tin in
        multiple structures using a variety
        of density functionals and assess
        their accuracy."
}
\end{lstlisting}
    \end{minipage} & 
    \cellcolor{lightgreen}\begin{minipage}[t]{4.45cm}\vspace{-10pt}
\begin{lstlisting}[style=json]
{
  "overarching_solution_domain":
        "High-throughput computational
        methods and density functional
        theory.",
  "solution_approach":
        "Using the high-throughput
        Automatic-FLOW (AFLOW) method
        to study tin's energetics with
        various density functionals.",
  "novelty_of_the_solution":
        "Application of Hubbard U
        corrections to improve predictions
        of phase transition temperatures."
}
\end{lstlisting}
    \end{minipage} & 
    \cellcolor{lightblue}\begin{minipage}[t]{4.45cm}\vspace{-10pt}
\begin{lstlisting}[style=json]
{
  "findings/results":
        "No functional is all satisfactory, 
        but Hubbard U corrections can be chosen 
        to predict the correct phase
        transition temperature.",
  "potential_impact_of_the_results":
        "Improved accuracy in predicting
        phase transitions in materials
        with small energy differences."
}
\end{lstlisting}
    \end{minipage} \\
    \end{tabular}
    
    \begin{tabular}{@{}p{13.9cm}@{}}
    \cdashline{1-1} \noalign{\vskip 0.2em}
    \textcolor{darkred}{\textbf{Contribution -  Topic}}: \textit{'Convergence Testing in Computational Simulations', 'Density Functional Theory (DFT) Accuracy', 'High-Throughput Computational Methods', 'Hubbard U Corrections', 'Tin Phase Transitions'} \\
    \bottomrule
    \end{tabular}
    \caption{Examples of extracted \textit{problem/solution/result/topic} contributions from scientific paper abstracts (example 2 of 2).}
\end{table}

\newpage
\subsection{Distribution of Extracted Topics}
This section shows the distribution of various topics extracted from the papers based on frequency. This gives us an idea of what kind of topics were extracted.
\begin{figure}[ht]
    \centering
    \begin{subfigure}[b]{\textwidth}
        \centering
        \includegraphics[width=\textwidth, keepaspectratio]{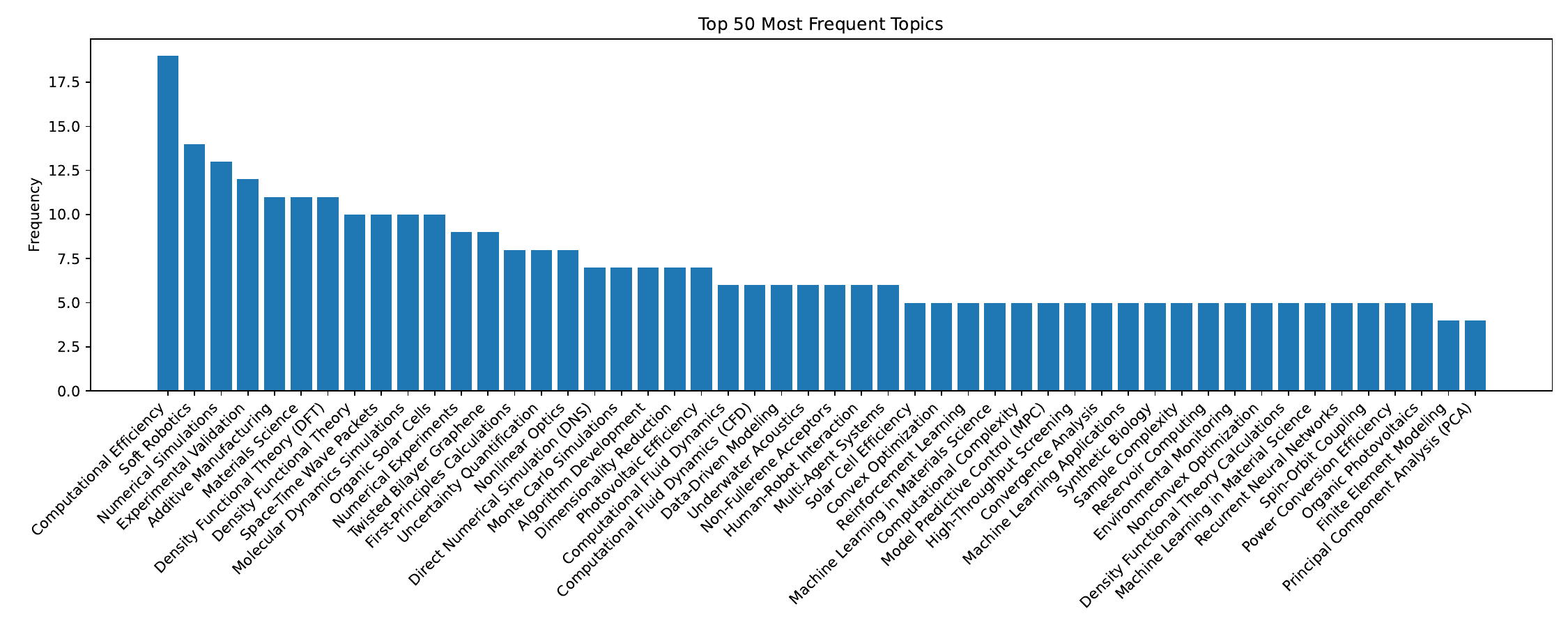}
        \caption{Top-50 topics by frequency in decreasing order}
    \end{subfigure}
    \vskip\baselineskip
    \begin{subfigure}[b]{\textwidth}
        \centering
        \includegraphics[width=\textwidth, keepaspectratio]{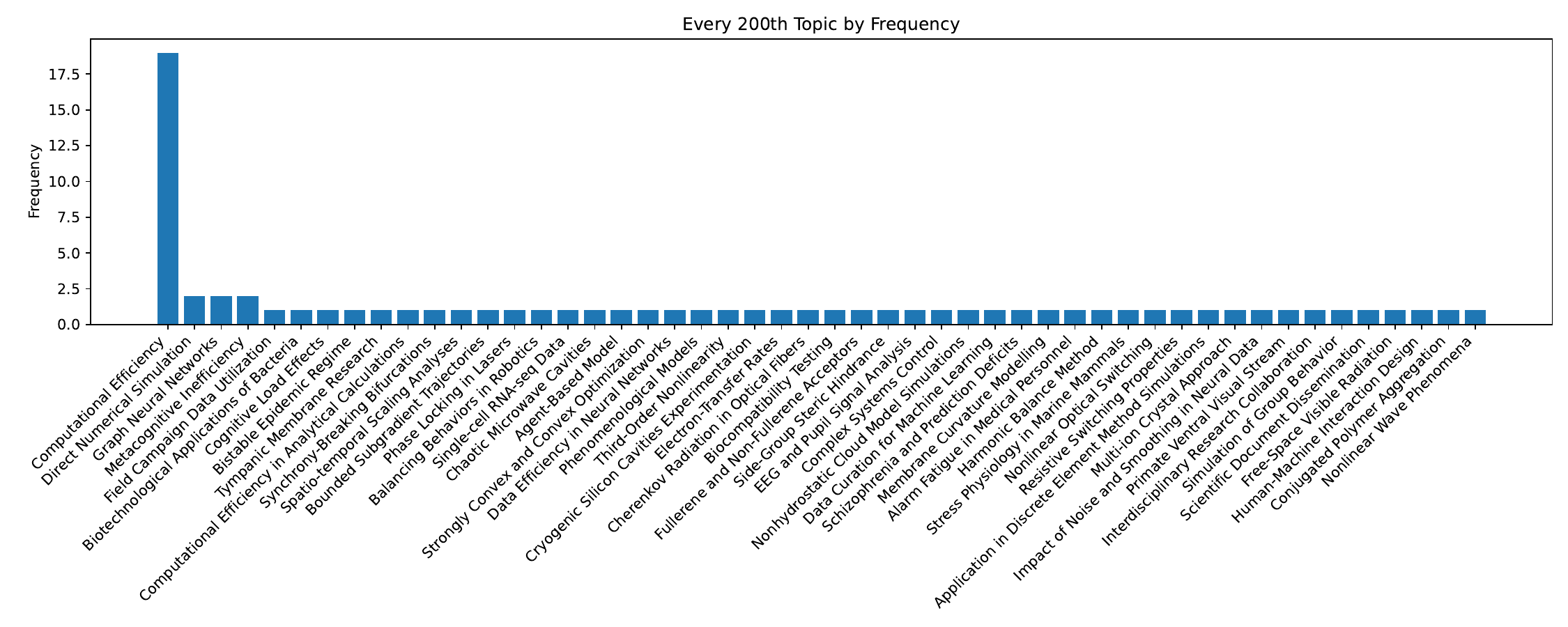}
        \caption{Sampled topics (every 200)}
    \end{subfigure}
    \caption{Distribution of topics extracted from \dataOne: (a) Top-50 topics, (b) Every 200 topics. Refer \S\ref{subsec:representing:papers} for more information.}
    \label{fig:topic-distribution}
\end{figure}

\clearpage

\section{Compiling a Seed Hierarchy}
\label{appendix:seed:hierarchy}

As we discuss in \S\ref{subsec:pure-llm:baselines}, we make a few adjustments to the seed hierarchy that we obtain from Wikipedia. Specifically: 

\begin{enumerate}
    \item We added ``Theoretical Computer Science'' and ``Information Theory'' as separate branches under ``Formal Sciences'' due to their unique characteristics;
    \item We moved ``Astronomy'' under ``Physical Science''; 
    \item ``Astronomy'', ``Geology'' and ``Oceanography'' are listed under ``Earth Science'' but since these topics are not specific to early, we move them up one layer so that they're directly under ``Physical Science''; 
The Wikipedia article groups Geology, Oceanography, and Meteorology under  ;
    \item We added ``History'' as a branch under ``Social Sciences''; 
    \item  We included ``Cell Biology'' and ``Genetics'' under ``Biological Sciences'' as they were relevant and their inclusion would only help in better hierarchy creation.
\end{enumerate}

These modifications aim to refine the hierarchy, ensuring it accurately reflects the distinct areas within each scientific domain.
The resulting hierarchy is included in Fig.\ref{fig:seed-hierarchy}.

\lstdefinestyle{mypython}{
    backgroundcolor=\color{backcolour},
    commentstyle=\color{codegreen},
    keywordstyle=\color{magenta},
    numberstyle=\tiny\color{codegray},
    stringstyle=\color{codepurple},
    basicstyle=\ttfamily\fontsize{7.5}{7.5}\selectfont,
    frame=single, 
    breakatwhitespace=false,         
    breaklines=true,                 
    captionpos=b,                    
    keepspaces=true,                 
    numbers=left,                    
    numbersep=5pt,                  
    showspaces=false,                
    showstringspaces=false,
    showtabs=false,                  
    tabsize=4
}

\begin{figure}[ht]
    \centering
    \begin{lstlisting}[language=XML, style=mypython]
{ 
    "Science":{
        "Formal Sciences":{
            "Logic":{},
            "Mathematics":{},
            "Statistics":{},
            "Computer Science":{},
            "Information Theory":{},
            "Systems Theory":{},
            "Decision Theory":{}
        },
        "Natural Sciences":{
            "Physical Science":{
                "Physics":{
                    "Classical Mechanics":{},
                    "Thermodynamics and statistical mechanics":{},
                    "Electromagnetism and photonics":{},
                    "Relativity":{},
                    "Quantum Mechanics":{},
                    "Atomic and molecular physics":{},
                    "Condensed matter physics":{},
                    "Optics and acoustics":{},
                    "High energy particle physics":{},
                    "Nuclear physics":{},
                    "Cosmology":{},
                    "Interdisciplinary Physics":{}
                },
                "Chemistry":{
                    "Physical Chemistry":{},
                    "Organic Chemistry":{},
                    "Inorganic Chemistry":{},
                    "Analytical Chemistry":{},
                    "Biological Chemistry":{},
                    "Theoretical Chemistry":{},
                    "Interdisciplinary Chemistry":{}
                },
                "Earth Science":{},
                "Astronomy":{},
                "Geology":{},
                "Oceanography":{},
                "Meteorology":{}
            },
            "Biological Sciences":{
                "Biochemistry":{},
                "Cell Biology":{},
                "Genetics":{},
                "Ecology":{},
                "Microbiology":{},
                "Botany":{},
                "Zoology":{}
            }
        },
        "Social Sciences":{
            "Anthropology":{},
            "Economics":{},
            "Political Science":{},
            "Sociology":{},
            "Psychology":{},
            "Geography":{},
            "History":{}
        }
    }
}
\end{lstlisting}
    \caption{The seed hierarchy used by our \baseline{} baselines. See  \S\ref{appendix:seed:hierarchy} for details. }
    \label{fig:seed-hierarchy}
\end{figure}

\clearpage

\section{\baseline: LLM-based Baselines}
This section includes the pipeline and prompts used for \baseline{} (parallel) and \baseline (incremental).
\label{appendix:baselines}
\subsection{Pipeline for \baselineOne}
This section demonstrates the pipeline used for \baseline{} (par) right from extracting topics and rationales to obtaining a final taxonomy with papers. 
\begin{figure}[htbp]
    \centering
    \includegraphics[width=1\linewidth,trim=0cm 0.35cm 0cm 0cm]{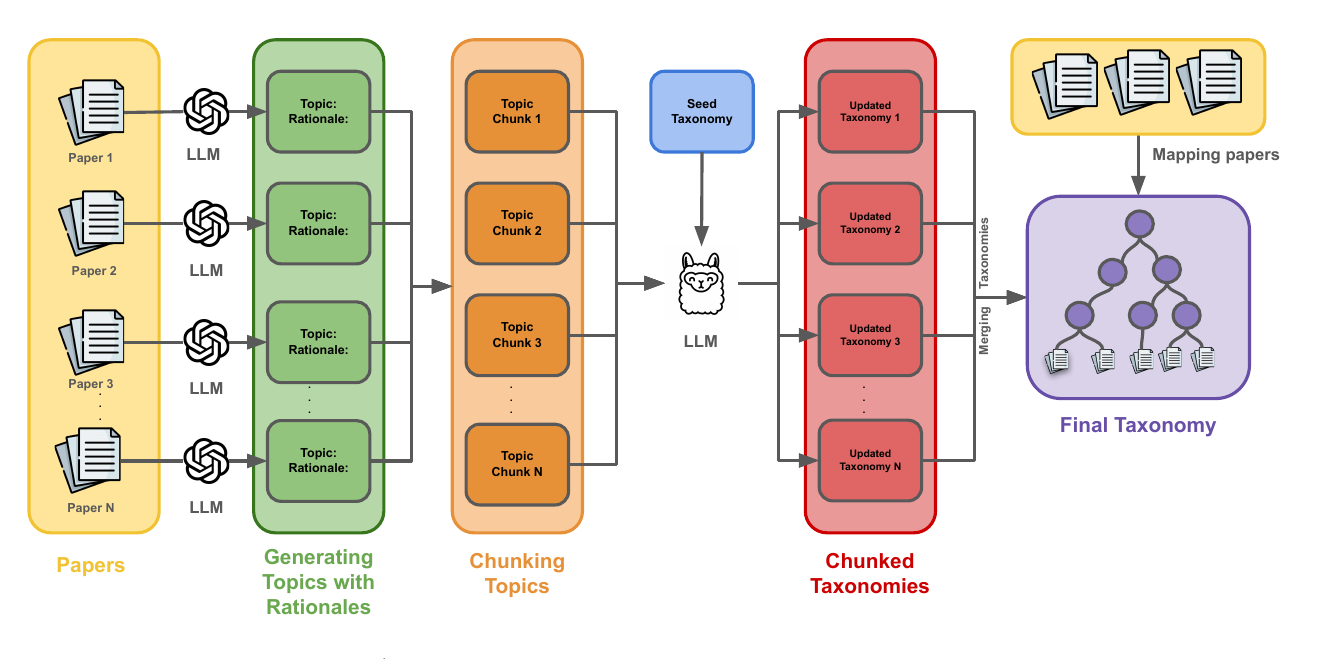}
    \caption{Pipeline for \baselineOne.}
    \label{fig:prompr-flmsci-par}
    
\end{figure}

\clearpage
\subsection{Prompt for \baselineOne}
This prompt guides a large language model (LLM) to expand an existing scientific taxonomy - the seed taxonomy (Refer to \ref{appendix:seed:hierarchy}) by adding a set of new topics in a logical and consistent manner. With a clear list of instructions it ensures accurate placement and also preserves the original structure. This prompt was used with \texttt{Llama-3.3-70B-Instruct}. 
\begin{figure}[htbp]
    \centering
    \includegraphics[width=1\linewidth,trim=0cm 0.55cm 0cm 0cm]{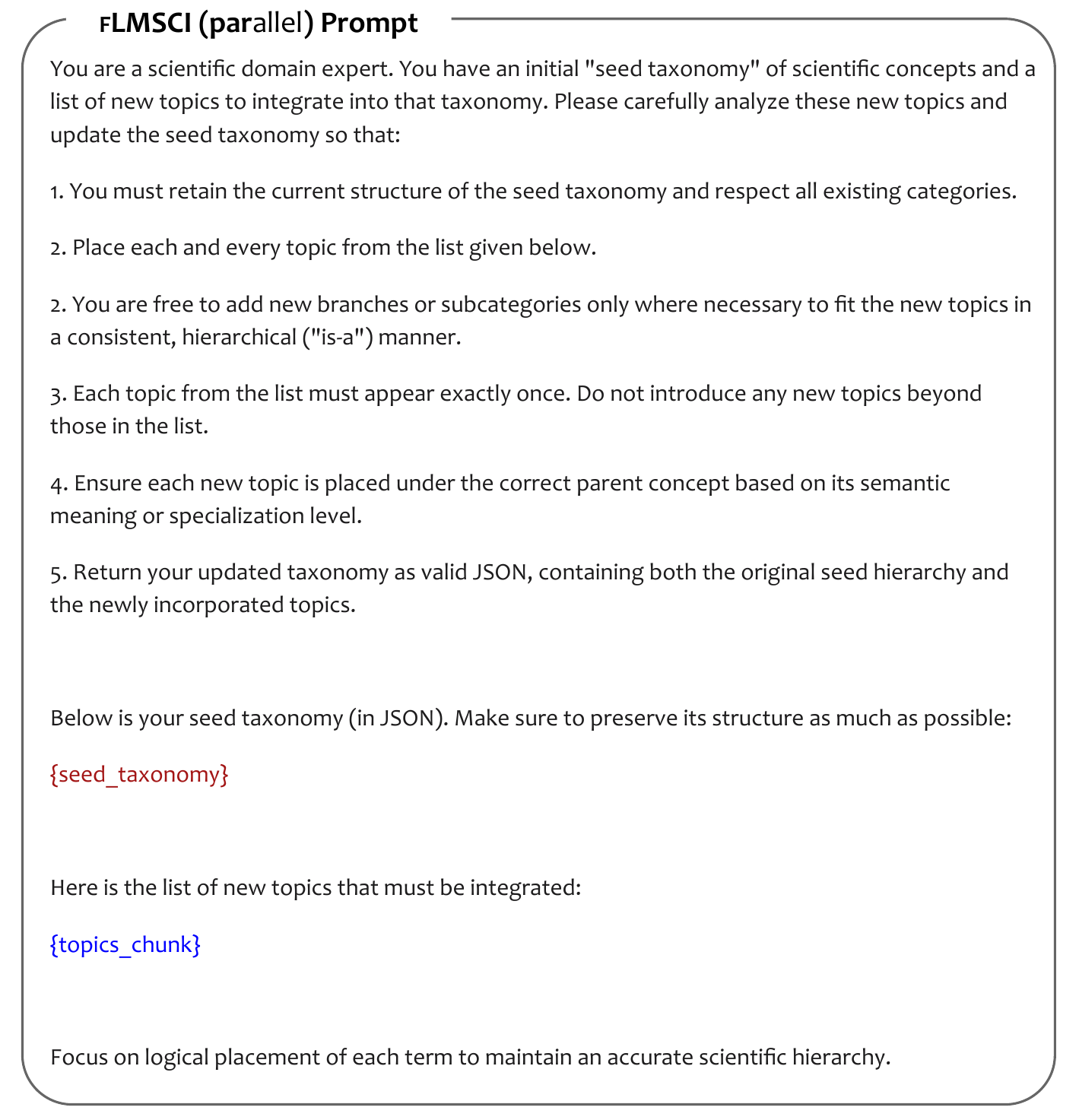}
    \caption{Prompt of \baselineOne{} pipeline}
    \label{fig:prompr-flmsci-incremental}
\end{figure}
\newpage

\subsection{Demonstration of actions for \baselineTwo{}}
This section demonstrates the various actions (\texttt{add sibling}, \texttt{make parent}, \texttt{go down} and \texttt{discard}) that are available for the LLM to take at various levels of taxonomy building. 
\begin{figure}[htbp]
    \centering
    \includegraphics[width=1\linewidth,trim=0cm 0.6cm 0cm 0cm]{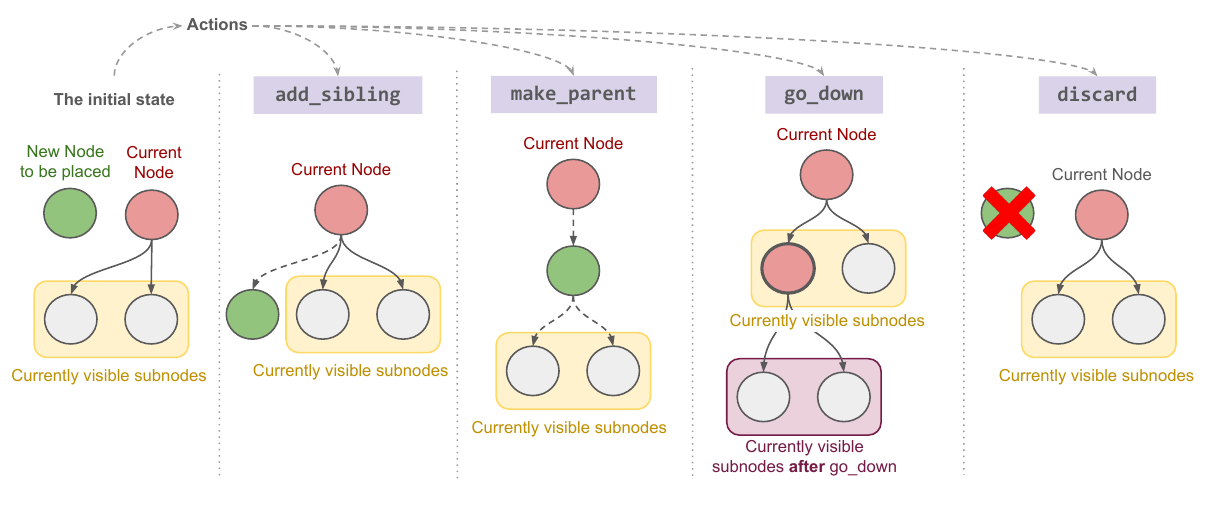}
    \caption{Actions for \baselineTwo}
    \label{fig:incremental:actions}
\end{figure}
\subsection{Prompt for \baselineTwo{}}
This prompt is used to place new scientific topics into an existing seed taxonomy (Refer \S\ref{appendix:seed:hierarchy}) incrementally. The model evaluates multiple possible actions based on the available action options. The prompt clearly instructs its priorities explicitly to give a hint to the model. The example usage and example output format help to get the response in a valid format. This prompt was used for \texttt{Llama-3.3-70B-Instruct}.

\begin{figure}[ht]
    \centering
\begin{lstlisting}[language=XML, style=mypython]
SUBNODE_DESCRIPTIONS = {
  "Formal Sciences": "Focuses on abstract systems and formal methodologies grounded in logic, mathematics, and symbolic reasoning. Provides theoretical frameworks (e.g., statistics, computer science, systems theory) used to model and solve problems across empirical disciplines and technology.",
  "Natural Sciences": "Investigates the physical universe and living organisms through empirical observation, experimentation, and theoretical analysis. Includes physical sciences (e.g., physics, chemistry, astronomy) and biological sciences (e.g., genetics, ecology) to uncover fundamental laws and processes governing nature.",
  "Social Sciences": "Studies human behavior, societies, and institutions using qualitative and quantitative methods. Encompasses disciplines like psychology, economics, and political science to analyze cultural, economic, and social interactions within historical and geographic contexts."
}
\end{lstlisting}
    \caption{  
    Descriptive statement used for contextualizing layer-1 items in the seed hierarchy, used in 
    \baselineTwo. 
    }
    \label{fig:descriptive-definitions}
\end{figure}

\begin{figure}[ht]
    \centering
    \begin{lstlisting}[language=XML, style=mypython]
You are building a scientific topics based hierarchy.

Path traced until now: {current_path}  
Subnode options available at this level: 
subnodes = [{subnodes}]
New topic: "{new_topic}"  

Evaluate all possible actions listed below equally before choosing the most appropriate one.  
Choose the action that best preserves a logical hierarchy, semantic clarity, and appropriate abstraction level.

**Priority Guidance**:
1. FIRST consider "go_down" if ANY existing subnode could reasonably contain the new topic as a specialization
2. THEN consider "make_parent" if multiple existing subnodes could be grouped under a new category
3. ONLY use "add_sibling" if the topic is FUNDAMENTALLY distinct from all existing subnodes at this level
4. "discard" should be used for low-quality or redundant topics

**Critical Rules**:
- A node about "Applications of X" should ALWAYS go under X, not as a sibling
- Specific methods/tools belong under their parent field (e.g., "PCR" under "Molecular Biology")
- Avoid creating flat structures

Possible actions:
1) "go_down" - Use this if the topic: {new_topic} is a *more specific* subtype of one of the "subnodes" and belongs *within* it.
2) "add_sibling" - Use this if the topic: {new_topic} is on the same level of abstraction as the existing "subnodes". It should be added *alongside* them as a direct child of `{current_path[-1]}`.
3) "discard" - Use this if the topic: {new_topic} is irrelevant, redundant, or already captured under another topic.
4) "make_parent" - Use this when the new topic: {new_topic} or any one of the "subnodes" is broader or more abstract than one or more of the subnodes. In that case, make the new topic a direct child of `{current_path[-1]}` and move the relevant subnodes under it. Return them in `"child_nodes": [...]`.

### Example of desired usage for each action:
1) "go_down"
   - "node": must be the name of one of the existing "subnodes"
   - "explanation": an optional text with reasoning
   - "child_nodes", "parent_node": not used.

2) "add_sibling"
   - "node": {new_topic}
   - "parent_node": {current_path[-1]}
   - "explanation": optional
   - "child_nodes": not used.

3) "discard"
   - "node": {new_topic}
   - "explanation": optional
   - "parent_node", "child_nodes": not used

4) "make_parent"
   - "node": {new_topic} or one of the "subnodes" whichever is more appropriate
   - "child_nodes": array of the subnodes that must be moved under the new node
   - "explanation": optional
   - "parent_node": not used

Your output must be valid JSON only:
{{
  "action": "go_down"|"add_sibling"|"make_parent"|"discard",
  "node": "string",
  "parent_node": "string or null",  // only used if action = add_sibling
  "child_nodes": ["string", ...],   // only used if action = make_parent
  "explanation": "string (optional)"
}}
No extra text.

\end{lstlisting}
    
    \caption{
    The detailed prompt used in the execution of our  
    \baselineTwo{} baseline. 
    }
    \label{fig:incremental:addition}
\end{figure}


\clearpage

\section{Further Details on Collection of Science Papers}
\label{appendix:collecting:papers}
\begin{figure}[htbp]
    \centering
    \includegraphics[width=1\linewidth,trim=0cm 0.8cm 0cm 0cm]{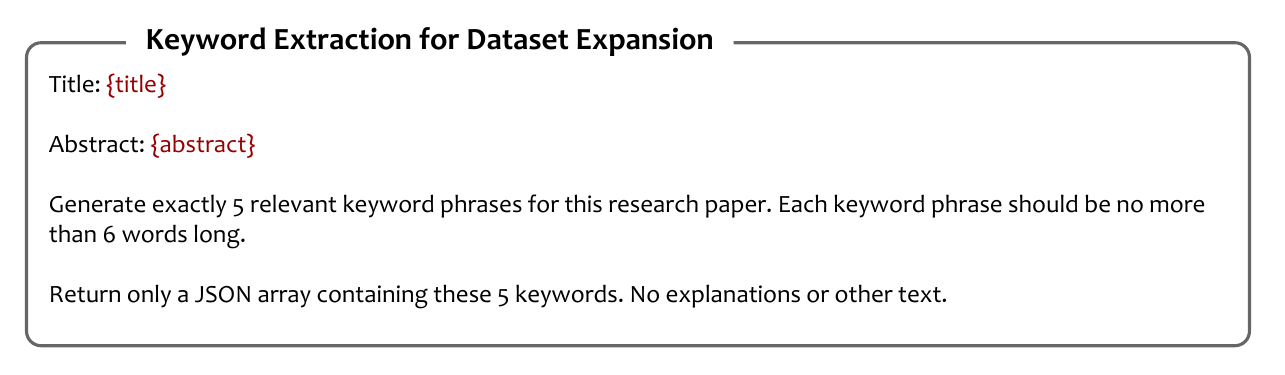}
    \caption{Prompt of Keyword Extraction for Dataset Expansion}
    \label{fig:prompt-dataset}
\end{figure}

This section provides more context on our piles of papers in our experiments from \S\ref{subsec:collecting:papers}. 
\dataTwo{} is an extension of \dataOne. For each paper in \dataOne, we extract five relevant keywords using an LLM (see Fig.\ref{fig:prompt-dataset}) and query the Semantic Scholar API\footnote{\url{https://www.semanticscholar.org/product/api}} with these keywords to retrieve additional relevant papers.

We apply three filtering criteria to ensure quality:
(a) \textbf{Citation Count:} A paper must have a minimum number of citations to be considered reliable. The minimum citation count is calculated using the formula: $(2 + 3 \times (2025 - \texttt{publish\_year})$.
(b) \textbf{Abstract Length:} A paper must have an abstract with at least 250 tokens, as measured by the tokenizer of \texttt{Llama-3.1-8B-Instruct}.
(c) \textbf{Publication Venue:} A paper must be published in a recognized journal or conference.
For each keyword, we select up to five papers that meet all criteria. 
This approach 
expands our seed corpus to $10K$ papers.

\section{Hyperparameters of \nameAlg{}}
\label{appendix:hyperparams}

Here shows the models and hyperparameters we use for the experiments mentioned in \S\ref{sec:hyperparams}. We utilize the GPT-4o model (\texttt{gpt-4o-2024-08-06}) to generate all contribution extractions along with detailed rationales explaining the extraction decisions. For \summarizer{}, we use  \texttt{Llama-3.3-70B-Instruct}~\citep{grattafiori2024llama3herdmodels} for its superiority of following instructions among open-source models, and use \texttt{gte-Qwen2-7B-instruct} as our \embedder{}. For clustering algorithm, we apply k-means clustering. The number of clusters for each layer is (6, 40, 276) when conducting experiments on \dataOne{} ($2K$ papers), and (6, 40, 276, 1250) when on \dataTwo{} ($10k$ papers).

\newpage
\section{Additional Experiments of \nameAlg{}}
\label{appendix:extra-exp}

\subsection{Evaluation Results on \dataOne{}}
\begin{table*}[!thbp]
\setlength{\tabcolsep}{2pt}
\centering
\footnotesize
\definecolor{lightblue}{RGB}{235,244,250}
\definecolor{lightgreen}{RGB}{237,246,240}
\definecolor{lightpurple}{RGB}{242,240,247}
\renewcommand{\arraystretch}{0.85}
\resizebox{\linewidth}{!}{%
\begin{tabular}{@{}l*{7}{c}}
\toprule
\multirow{2}{*}{Method} & \multicolumn{2}{c}{Accuracy (\%)} & \multicolumn{2}{c}{LLM Cost} & \multicolumn{3}{c}{Hierarchy Structure} \\
\cmidrule(lr){2-3} \cmidrule(lr){4-5} \cmidrule(lr){6-8}
& Strict-Acc $\uparrow$ & L1-Acc $\uparrow$ & \specialcell{Avg. \# of\\Input Tokens $\downarrow$} & \specialcell{\# of\\Calls $\downarrow$} & Depth & \specialcell{Avg. Branching\\Factor} & \specialcell{Max. Branching\\Factor} \\
\midrule
\rowcolor{lightpurple} \multicolumn{8}{@{}l}{\textit{Contributions type: Problem Statement}} \\
\midrule
\nameAlg{}  & {\textbf{51.1 \smallfont{$\pm$ 3.8}}} & {\textbf{81.7 \smallfont{$\pm$ 2.6}}} & 2624 & & & & 20 \\
\arrowdownright  Top-down & 49.0 \smallfont{$\pm$ 3.7} & 80.3 \smallfont{$\pm$ 2.7} & 2953 & 322 & 3 & 7.1  & 18 \\
\arrowdownright  Bottom-up & 45.9 \smallfont{$\pm$ 5.0} & 69.3 \smallfont{$\pm$ 8.1} & 2177 & & & & 16 \\
\midrule
\rowcolor{lightgreen} \multicolumn{8}{@{}l}{\textit{Contributions type: Solution Statement}} \\
\midrule
\nameAlg{}  & {\textbf{48.8 \smallfont{$\pm$ 6.1}}} & {\textbf{82.3 \smallfont{$\pm$ 1.1}}} & 2343 & & & & 16 \\
\arrowdownright Top-down & 45.9 \smallfont{$\pm$ 5.5} & 79.2 \smallfont{$\pm$ 3.4} & 2521 & 322 & 3 & 7.1 & 19 \\
\arrowdownright Bottom-up & 36.7 \smallfont{$\pm$ 2.6} & 67.0 \smallfont{$\pm$ 4.3} & 1990 & & & & 14 \\
\midrule
\rowcolor{lightblue} \multicolumn{8}{@{}l}{\textit{Contributions type: Results Statement}} \\
\midrule
\nameAlg{} & 46.4 \smallfont{$\pm$ 5.2} & 76.4 \smallfont{$\pm$ 6.9} & 2654 & & & & 16 \\
\arrowdownright Top-down & \textbf{47.3 \smallfont{$\pm$ 3.1}} & \textbf{80.5 \smallfont{$\pm$ 4.4}} & 2718 & 322 & 3 & 7.1 & 16 \\
\arrowdownright Bottom-up & 40.0 \smallfont{$\pm$ 10.7} & 64.0 \smallfont{$\pm$ 8.9} & 2210 & & & & 13 \\
\bottomrule
\end{tabular}}%
\renewcommand{\arraystretch}{1}
\caption{
    Evaluation results of \nameAlg{} and the corresponding baselines on the 2K (\dataOne{}) dataset. 
}
\label{tab:exp-results-small}
\end{table*}

\subsection{Evaluation Results on \dataTwo{}}

\begin{table*}[!thbp]
\setlength{\tabcolsep}{2pt}
\centering
\footnotesize
\definecolor{lightblue}{RGB}{235,244,250}
\definecolor{lightgreen}{RGB}{237,246,240}
\definecolor{lightpurple}{RGB}{242,240,247}
\renewcommand{\arraystretch}{0.85}
\resizebox{\linewidth}{!}{%
\begin{tabular}{@{}l*{7}{c}}
\toprule
\multirow{2}{*}{Method} & \multicolumn{2}{c}{Accuracy (\%)} & \multicolumn{2}{c}{LLM Cost} & \multicolumn{3}{c}{Hierarchy Structure} \\
\cmidrule(lr){2-3} \cmidrule(lr){4-5} \cmidrule(lr){6-8}
& Strict-Acc $\uparrow$ & L1-Acc $\uparrow$ & \specialcell{Avg. \# of\\Input Tokens $\downarrow$} & \specialcell{\# of\\Calls $\downarrow$} & Depth & \specialcell{Avg. Branching\\Factor} & \specialcell{Max. Branching\\Factor} \\
\midrule
\rowcolor{lightpurple} \multicolumn{8}{@{}l}{\textit{Contributions type: Problem Statement}} \\
\midrule
\nameAlg{} & {\textbf{43.7} \smallfont{$\pm$ 6.5}} & 85.8 \smallfont{$\pm$ 4.2} &  7451 & & & & 26 \\
\arrowdownright  Top-down & 41.5 \smallfont{$\pm$ 8.2 } &  {\textbf{86.5} \smallfont{$\pm$ 5.6}} & 8990 & 1572 & 4 & 8 & 30 \\
\arrowdownright  Bottom-up & 26.2 \smallfont{$\pm$ 5.4} &  41.9 \smallfont{$\pm$ 4.0} & 5924 & & & & 26 \\
\midrule
\rowcolor{lightgreen} \multicolumn{8}{@{}l}{\textit{Contributions type: Solution Statement}} \\
\midrule
\nameAlg{} & {\textbf{24.7} \smallfont{$\pm$ 4.8}} & {\textbf{65.8} \smallfont{$\pm$ 2.5}} &  7653 & & & & 28 \\
\arrowdownright  Top-down & 22.4 \smallfont{$\pm$ 3.5 } &  52.3 \smallfont{$\pm$ 3.0} & 4032 & 1572 & 4 & 8 & 26 \\
\arrowdownright  Bottom-up & 23.9 \smallfont{$\pm$ 3.3} &  51.3 \smallfont{$\pm$ 3.1} & 6150 & & & & 28 \\
\midrule
\rowcolor{lightblue} \multicolumn{8}{@{}l}{\textit{Contributions type: Results Statement}} \\
\midrule
\nameAlg{} & {\textbf{27.6} \smallfont{$\pm$ 4.6}} & {\textbf{69.8} \smallfont{$\pm$ 2.1}} &  6457 & & & & 30 \\
\arrowdownright  Top-down & 19.7 \smallfont{$\pm$ 4.0 } &  54.0 \smallfont{$\pm$ 3.3} & 5380 & 1572 & 4 & 8 & 30 \\
\arrowdownright  Bottom-up & 23.6 \smallfont{$\pm$ 2.7} &  55.2 \smallfont{$\pm$ 2.9} & 4731 & & & & 28 \\
\bottomrule
\end{tabular}}%
\renewcommand{\arraystretch}{1}
\vspace{-0.05in}
\caption{
    Evaluation results of \nameAlg{} and the corresponding baselines on the 10K (\dataTwo{}) dataset. \nameAlg{} maintains high accuracy across all contribution types, proving the rationale behind our hybrid design. The \textit{problem statement} contribution type consistently yields the most accurate hierarchies, indicating this contribution type contributes most to hierarchy construction. 
}
\vspace{-1em}
\label{tab:exp-results-large}
\end{table*}

\subsection{Detailed Evaluation Results on \textit{Topic} Contributions}
\label{appendix:topic_result}
Here we show the complete evaluation results mentioned in \S\ref{subsec:evaluation:utiliization}. \nameAlg{}, \baseline{} (\textbf{par}allel) and \baseline (\textbf{inc}remental) are using \textit{Topic} as contribution type.
\begin{table*}[!thbp]
\setlength{\tabcolsep}{0.5pt}
  \centering
  \small 
  \resizebox{\textwidth}{!}{
  \begin{tabular}{@{}l*{7}{c}@{}}
    \toprule
    \multirow{2}{*}{Method} & \multicolumn{2}{c}{Accuracy (\%)} & \multicolumn{2}{c}{LLM Cost} & \multicolumn{3}{c}{Hierarchy Structure} \\
    \cmidrule(lr){2-3} \cmidrule(lr){4-5} \cmidrule(lr){6-8}
    & Strict-Acc $\uparrow$ & L1-Acc $\uparrow$ & \specialcell{Avg. \# of \\Input Tokens $\downarrow$} & \specialcell{\# of\\Calls $\downarrow$} & \specialcell{Avg Bran.\\Factor} & \specialcell{Max Bran.\\Factor} & \specialcell{\# of\\Items} \\
    \midrule
    \rowcolor{lightblue} \multicolumn{8}{@{}l}{\textit{Contributions type: Topic}} \\
    \midrule
    \nameAlg{} & {\textbf{14.9} \smallfont{$\pm$ 2.7}} & {\textbf{65.7} \smallfont{$\pm$ 4.4}} & 5017 & 322 & 40.9 & 128 & 11k \\
    \arrowdownright  Top-down & 14.5 \smallfont{$\pm$ 4.7} & 62.5 \smallfont{$\pm$ 7.4} & 6440 & 322 & 40.9 & 104 & 11k \\
    \arrowdownright  Bottom-up & 13.9 \smallfont{$\pm$ 5.3} & 54.4 \smallfont{$\pm$ 12.7} & 3988 & 322 & 40.9 & 119 & 11k \\
    \graymidrule
    \arrowdownright \baseline{} (par) & 4.0 \smallfont{$\pm$ 2.8} & 32.0 \smallfont{$\pm$ 6.3} & 8896 & 226 & 13.9 & 734 & {\color{highlight}\textbf{9.9K}} \\
    \arrowdownright   \baseline{} (inc) & {\textbf{18.0 \smallfont{$\pm$ 5.3}}} & {\textbf{91.0 \smallfont{$\pm$ 4.0}}} & 4040 & {\color{highlight}\textbf{61K}} & 3.6 & 704 & {\color{highlight}\textbf{10.4K}} \\
    \bottomrule
  \end{tabular}
  }
  \caption{
  Evaluation results of \nameAlg{}, \baseline{} (\textbf{par}allel) and \baseline (\textbf{inc}remental) when using \textit{Topic} as contribution type.
  ``Bran.'' stands for ``Branching''. 
  All methods show poor Strict-Acc ($\leq 18.0\%$), highlighting the challenging nature of the task. 
  On one hand,  \baseline{} (inc) achieves the highest accuracy, showing the feasibility of building hierarchies by pure LLM-based methods. 
  However, it incurs substantial computational costs, about 190$\times$ higher than other methods. In contrast, \nameAlg{} offers a balanced performance profile with competitive accuracy (14.9\% Strict-Acc, 65.7\% L1-Acc) while maintaining significantly lower computational requirements. 
  }
  \label{tab:topics-2k-detail}
\end{table*}

\subsection{Comparison of Different Embedding models}
\label{appendix:embedding}
For the \embedder ~mentioned in \S\ref{sec:muhan:approach}. We evaluate three embedding models—Qwen’s \texttt{gte-Qwen2-7B-instruct}~\citep{li2023towards}, OpenAI’s \texttt{text-embedding-3-large}, and \texttt{text-embedding-ada-002}. The first two perform similarly, whereas \texttt{text-embedding-ada-002} produces markedly weaker results. Given the comparable overall performance between the two leading models, we select \texttt{gte-Qwen2-7B-instruct} for our main experiments due to its strong balanced performance across both metrics, superior Strict-Acc results, and practical advantages as an open-weight model that offers greater accessibility and cost-effectiveness for reproducible research. 
\begin{table}[th]
\setlength{\tabcolsep}{3pt}
  \centering
  \small 
  \begin{tabular}{@{}l*{6}{c}@{}}
    \toprule
    Models→ & \multicolumn{2}{c}{\texttt{text-embedding-3-large}} & \multicolumn{2}{c}{\texttt{gte-Qwen2-7B-instruct}}  & \multicolumn{2}{c}{\texttt{text-embedding-ada-002}}  \\
    \cmidrule(lr){2-3} \cmidrule(lr){4-5} \cmidrule(lr){6-7}
    Metrics→ & L1-Acc & Sctric-Acc & L1-Acc & Sctric-Acc & L1-Acc & Sctric-Acc \\
    \midrule
    PROBLEM  & {\textbf{86.7 \smallfont{$\pm$ 4.6}}} & 46.7 \smallfont{$\pm$ 0.9} & 81.7 \smallfont{$\pm$ 2.6} & {\textbf{51.1 \smallfont{$\pm$ 3.8}}} & 76.0 \smallfont{$\pm$ 4.4} & 41.7 \smallfont{$\pm$ 5.2} \\
    SOLUTION & 80.3 \smallfont{$\pm$ 3.4} & 36.7 \smallfont{$\pm$ 1.7} & {\textbf{82.3 \smallfont{$\pm$ 1.1}}} & {\textbf{48.8 \smallfont{$\pm$ 6.1}}} & 63.5 \smallfont{$\pm$ 2.0} & 31.0 \smallfont{$\pm$ 3.2} \\
    RESULTS & {\textbf{84.7 \smallfont{$\pm$ 5.7}}} & 44.0 \smallfont{$\pm$ 0.8} & 76.4 \smallfont{$\pm$ 6.9} & {\textbf{46.4 \smallfont{$\pm$ 5.2}}} & 74.6 \smallfont{$\pm$ 3.4} & 41.0 \smallfont{$\pm$ 8.7} \\
    \bottomrule
  \end{tabular}
  \caption{Performance comparison across three embedding models and contribution types. \texttt{gte-Qwen2-7B-instruct} demonstrates superior Sctric-Acc performance across all categories, while \texttt{text-embedding-3-large} excels in L1-Acc for \textit{problem} and \textit{results}. \texttt{text-embedding-ada-002} shows consistently weaker performance across both metrics.}
  \label{tab:embedding_comparison}
\end{table}

\subsection{Experiments of Prompt Engineering}
\label{appendix:prompt}
We investigate the effect of different prompts on the final quality of hierarchy. In the main text, for the \summarizer ~mentioned in \S\ref{sec:muhan:approach}, we use the detailed version prompt which is carefully curated. For comparison, we also conduct the experiments with a much simpler prompt. 
\begin{table}[!htbp]
\centering
\footnotesize
\renewcommand{\arraystretch}{1.2}  
\setlength{\tabcolsep}{5pt}
\label{tab:prompt_comparison_text}
\begin{tabular}{|p{6.6cm}|p{6.6cm}|}
\hline
\textbf{Detailed (Curated) Prompt} & \textbf{Simple Prompt} \\[0.3ex]
\hline
\rule{0pt}{2.5ex}You are a scientific research expert specializing in identifying and analyzing research problems and challenges. Your task is to analyze a collection of research papers or research clusters and provide a focused analysis of the research problems they address.

The input could be either a collection of individual papers or research cluster summaries. Based on the content, you need to:
\begin{enumerate}
\setlength{\itemsep}{0pt}
\setlength{\parskip}{0pt}
\item Identify the core research problems and challenges being addressed
\item Determine the overarching problem domain that connects these research efforts
\item Analyze the specific difficulties, gaps, or limitations that motivate this research
\item Understand the research questions or goals that drive these investigations
\item Generate an appropriate cluster name that captures the essence of the problem space
\item Provide a comprehensive problem-focused analysis
\end{enumerate}

Here is the content to analyze: \{\}

Remember to:
\begin{itemize}
\setlength{\itemsep}{0pt}
\setlength{\parskip}{0pt}
\item Focus specifically on problems, challenges, and research gaps rather than solutions
\item Be specific about the technical difficulties and limitations being addressed
\item Identify both theoretical and practical challenges
\item Consider interdisciplinary connections in problem formulation
\item Maintain scientific accuracy and use precise terminology
\item Generate only one JSON format output that must follow the structure exactly
\end{itemize}

Please format your response as a JSON object with the following structure:
\begin{lstlisting}[basicstyle=\tiny\ttfamily, breaklines=true, columns=fullflexible]
{
    "Cluster Name": "A clear and specific title focusing
    on the problem domain (No less than 5 words)",
    
    "Problem": {
        "overarching problem domain": "The broad scientific 
        domain where these problems exist",
        "challenges/difficulties": "Specific technical,
        theoretical, or practical challenges that these 
        papers address",
        "research question/goal": "The fundamental 
        research questions or objectives that motivate
        this work"
    }
}
\end{lstlisting}
\rule[-1.5ex]{0pt}{2ex}
& 
\rule{0pt}{2.5ex}You are a scientific research expert specializing in identifying and analyzing research problems and challenges.

Analyze the input \%s and output one JSON object:
\begin{lstlisting}[basicstyle=\tiny\ttfamily, breaklines=true, columns=fullflexible]
{
    "Cluster Name": "A clear and specific title (No less 
    than 5 words)",
    "Problem": {
    "overarching problem domain": "",
    "challenges/difficulties": "",
    "research question/goal": ""
  }
}
\end{lstlisting}

\textbf{Instructions}

Extract key themes and concepts.

Identify the common thread that links the items.

Craft a clear, specific title ($\geq$ 5 words) for Cluster Name.

Return only the JSON—nothing else.
\rule[-1.5ex]{0pt}{2ex}
\\
\hline
\end{tabular}
\caption{Comparison of Detailed (Curated) and Simple Prompts}
\end{table}

\newpage
The results show that across all contributions, the curated prompt offers significantly better quality hierarchies.
\begin{table}[thbp]
\setlength{\tabcolsep}{3pt}
  \centering
  \small 
  \begin{tabular}{@{}lc*{4}{c}@{}}
    \toprule
    \multirow{2}{*}{Prompt type ↓} & Embedder→ & \multicolumn{2}{c}{\texttt{text-embedding-3-large}} & \multicolumn{2}{c}{\texttt{gte-Qwen2-7B-instruct}}  \\
    \cmidrule(lr){3-4} \cmidrule(lr){5-6}
    & Metrics→ & L1-Acc & Sctric-Acc & L1-Acc & Sctric-Acc \\
    \midrule
    Simplified & \multirow{2}{*}{\textit{problem}} & 75.0 \smallfont{$\pm$ 4.6} & 33.7 \smallfont{$\pm$ 3.7} & 61.0 \smallfont{$\pm$ 0.8} & 24.7 \smallfont{$\pm$ 1.7} \\
    Detailed &  & {\textbf{86.7 \smallfont{$\pm$ 4.6}}} & {\textbf{46.7 \smallfont{$\pm$ 0.9}}} & {\textbf{81.7 \smallfont{$\pm$ 2.6}}} & {\textbf{51.1 \smallfont{$\pm$ 3.8}}} \\
    \midrule
    Simplified & \multirow{2}{*}{\textit{solution}} & 65.3 \smallfont{$\pm$ 3.4} & 32.7 \smallfont{$\pm$ 2.6} & 59.0 \smallfont{$\pm$ 2.8} & 21.7 \smallfont{$\pm$ 2.9} \\
    Detailed &  & {\textbf{80.3 \smallfont{$\pm$ 3.4}}} & {\textbf{36.7 \smallfont{$\pm$ 1.7}}} & {\textbf{82.3 \smallfont{$\pm$ 1.1}}} & {\textbf{48.8 \smallfont{$\pm$ 6.1}}} \\
    \midrule
    Simplified & \multirow{2}{*}{\textit{results}} & 77.7 \smallfont{$\pm$ 4.1} & 38.0 \smallfont{$\pm$ 4.6} & 66.7 \smallfont{$\pm$ 3.3} & 27.7 \smallfont{$\pm$ 2.5} \\
    Detailed &  & {\textbf{84.7 \smallfont{$\pm$ 5.7}}} & {\textbf{44.0 \smallfont{$\pm$ 0.8}}} & {\textbf{76.4 \smallfont{$\pm$ 6.9}}} & {\textbf{46.4 \smallfont{$\pm$ 5.2}}} \\
    \bottomrule
  \end{tabular}
  \caption{Performance comparison between simplified and detailed prompts across different embedding models and contribution types. Detailed prompts consistently outperform simplified prompts across all scenarios, with improvements ranging from 7.0 to 23.3 \% for L1-Acc and 3.0 to 26.4 \% for Sctric-Acc. The \texttt{gte-Qwen2-7B-instruct} model shows the largest performance gains, with L1-Acc improvements of 20.7, 23.3, and 9.7 \% for \textit{problem}, \textit{solution}, and \textit{results} respectively.}
  \label{tab:prompt_comparison_exp}
\end{table}

\newpage
\section{Visualization and Examples of Inter-Cluster Citations}
\label{appendix:citation}
Below is the figure of comparing inter- and intra-cluster citation counts mentioned in \S\ref{subsection:additional_ana}.
\begin{figure}[tbh]
    \centering
    \includegraphics[width=0.55\linewidth]{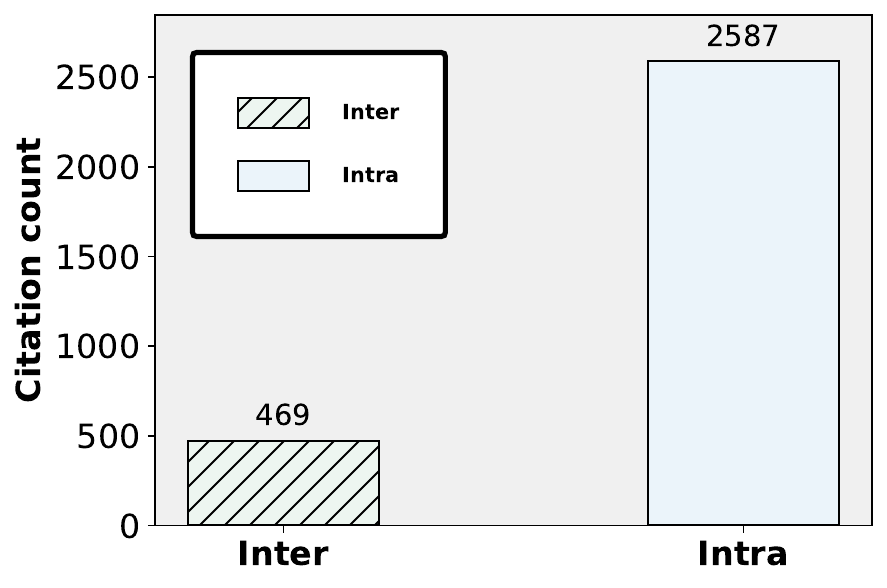}
    \caption{Comparison of inter(green)- and intra(blue)-cluster citation counts. 84.7\% citations are between papers in the same top layer cluster, and the rest inter-cluster citations are mostly theory-to-application works, which proves the reliability of \nameAlg{}.}
    \label{fig:citation_count}
\end{figure}

\newpage
Below are examples of citations between papers in different top-layer clusters. These examples show that many inter-cluster citations represent theory-to-application connections, while the last row illustrates cross-disciplinary citations between research fields. Importantly, all papers involved are correctly categorized—the inter-cluster citations reflect legitimate relationships rather than classification errors.
\begin{table}[h]
\centering
\small
\begin{tabular}{@{}p{0.46\textwidth}@{\hspace{2mm}}c@{\hspace{2mm}}p{0.48\textwidth}@{}}
\toprule
\textbf{Citing Paper} & & \textbf{Cited Paper} \\
\midrule
\multicolumn{3}{@{}l}{\small\textit{Rationale: AI research grounding in foundational cognitive science theory}} \\[1mm]
\hdashline[1pt/2pt]
\multicolumn{3}{c}{}\\[-2mm]
\parbox[t][3.2\baselineskip][b]{0.46\textwidth}{\raggedright
\textbf{Minding Language Models' (Lack of) Theory of Mind: A Plug-and-Play Multi-Character Belief Tracker}
\vfill {\footnotesize\textit{Challenges and Limitations in ML and AI}}}
& 
\raisebox{-1.2ex}{$\xrightarrow{}$} & 
\parbox[t][3.2\baselineskip][b]{0.48\textwidth}{\raggedright
\textbf{Does the chimpanzee have a theory of mind?}
\vfill {\footnotesize\textit{Neuroscience, Cognitive Psychology, and Neurotechnology}}}
\\[2mm]
\hline
\multicolumn{3}{c}{}\\[-2mm]
\multicolumn{3}{@{}l}{\small\textit{Rationale: Theory-to-application for THz photonics}} \\[1mm]
\hdashline[1pt/2pt]
\multicolumn{3}{c}{}\\[-2mm]
\parbox[t][3.2\baselineskip][b]{0.46\textwidth}{\raggedright
\textbf{Terahertz topological photonic integrated circuits for 6G and beyond}
\vfill {\footnotesize\textit{Advanced Materials Challenges}}}
& 
\raisebox{-1.2ex}{$\xrightarrow{}$} & 
\parbox[t][3.2\baselineskip][b]{0.48\textwidth}{\raggedright
\textbf{Topological photonics}
\vfill {\footnotesize\textit{Quantum Systems and Materials Science}}}
\\[2mm]
\hline
\multicolumn{3}{c}{}\\[-2mm]
\multicolumn{3}{@{}l}{\small\textit{Rationale: Hardware implementation citing quantum network theory}} \\[1mm]
\hdashline[1pt/2pt]
\multicolumn{3}{c}{}\\[-2mm]
\parbox[t][3.2\baselineskip][b]{0.46\textwidth}{\raggedright
\textbf{Cavity electro-optics in thin-film lithium niobate}
\vfill {\footnotesize\textit{Advanced Materials Challenges}}}
& 
\raisebox{-1.2ex}{$\xrightarrow{}$} & 
\parbox[t][3.2\baselineskip][b]{0.48\textwidth}{\raggedright
\textbf{Quantum internet: A vision for the road ahead}
\vfill {\footnotesize\textit{Quantum Systems and Materials Science}}}
\\[2mm]
\hline
\multicolumn{3}{c}{}\\[-2mm]
\multicolumn{3}{@{}l}{\small\textit{Rationale: Manufacturing citing characterization techniques}} \\[1mm]
\hdashline[1pt/2pt]
\multicolumn{3}{c}{}\\[-2mm]
\parbox[t][3.2\baselineskip][b]{0.46\textwidth}{\raggedright
\textbf{Creating Quantum Emitters in Hexagonal Boron Nitride}
\vfill {\footnotesize\textit{Advanced Materials Challenges}}}
& 
\raisebox{-1.2ex}{$\xrightarrow{}$} & 
\parbox[t][3.2\baselineskip][b]{0.48\textwidth}{\raggedright
\textbf{Nanoscale Imaging and Control of hBN Single Photon Emitters}
\vfill {\footnotesize\textit{Quantum Systems and Materials Science}}}
\\[2mm]
\hline
\multicolumn{3}{c}{}\\[-2mm]
\multicolumn{3}{@{}l}{\small\textit{Rationale: Cross-disciplinary bridge between biology and quantum physics}} \\[1mm]
\hdashline[1pt/2pt]
\multicolumn{3}{c}{}\\[-2mm]
\parbox[t][3.2\baselineskip][b]{0.46\textwidth}{\raggedright
\textbf{Magnetic field effects in biology from radical pair mechanism}
\vfill {\footnotesize\textit{Neuroscience, Cognitive Psychology, and Neurotechnology}}}
& 
\raisebox{-1.2ex}{$\xrightarrow{}$} & 
\parbox[t][3.2\baselineskip][b]{0.48\textwidth}{\raggedright
\textbf{Quantum biology revisited}
\vfill {\footnotesize\textit{Quantum Systems and Materials Science}}}
\\[3mm]
\bottomrule
\end{tabular}
\caption{Examples of cross-cluster citations. Each row shows the citing paper, the cited paper, their cluster names, and the citation rationale.}
\end{table}

\newpage
\section{Demonstration of Hierarchy}
\label{sec:visualization}
Below is a snippet of our final hierarchy result.
\begin{figure}[htbp]
    \centering
    \includegraphics[width=1\linewidth,trim=0cm 0.8cm 0cm 0cm]{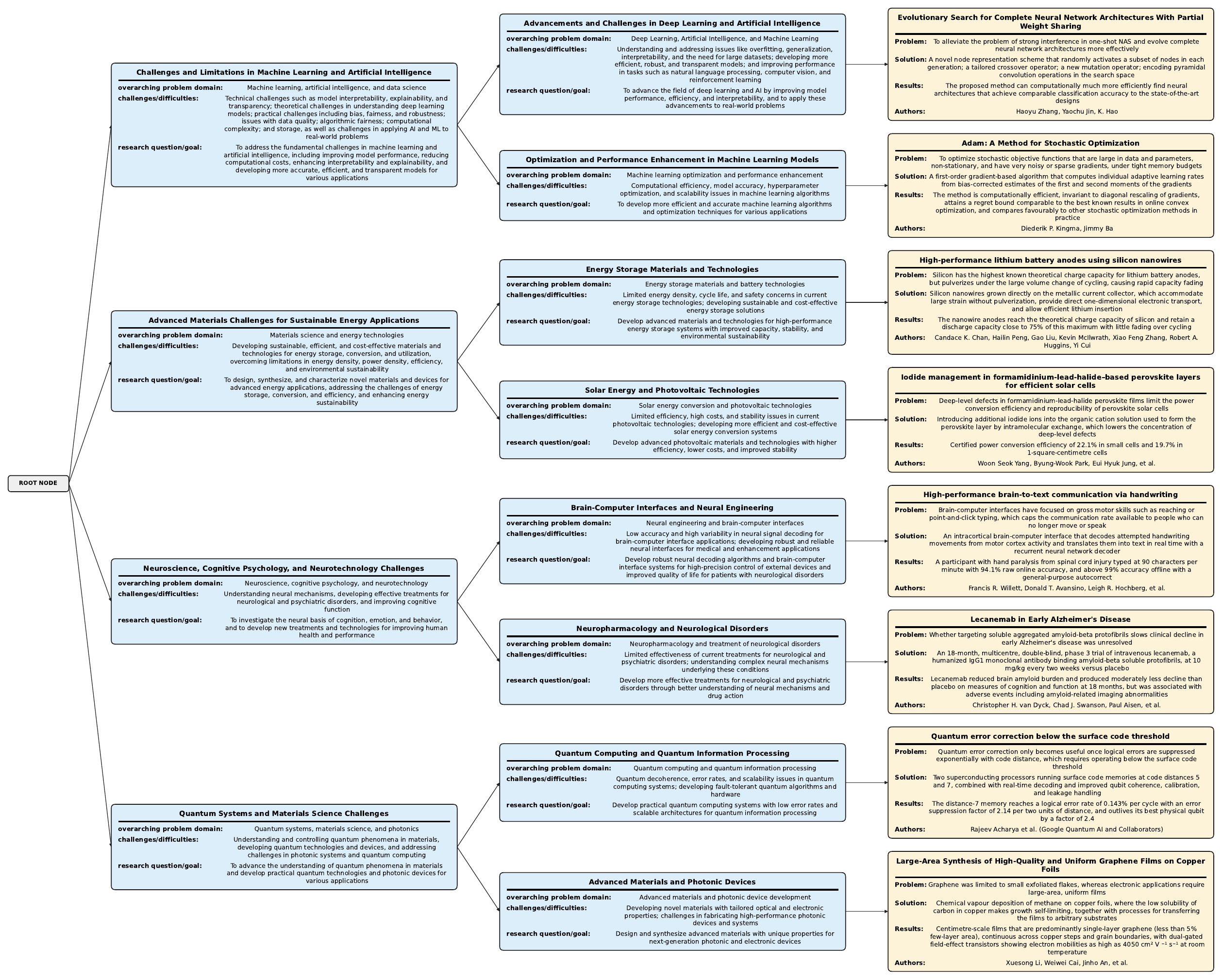}
    \caption{Above is a small example of a final hierarchy generated by \nameAlg{} on the \dataTwo~ dataset. The original hierarchy has 4 levels, use papers' \textit{problem} contribution. Due to space constraints, this snippet shows only two levels of clusters above the individual papers. }
    \label{fig:hierarchy-demon}
\end{figure}

\end{document}